\RequirePackage{iftex}
\ifPDFTeX\pdfoutput=1\fi

\documentclass{article}

\PassOptionsToPackage{numbers, sort&compress}{natbib}
\usepackage[main, final]{neurips_2026}

\usepackage[utf8]{inputenc} 
\usepackage[T1]{fontenc}    
\usepackage{hyperref}
\usepackage{url}            
\usepackage{booktabs}       
\usepackage{placeins}
\usepackage{longtable}      
\usepackage{amsfonts}       
\usepackage{nicefrac}       
\usepackage{microtype}      
\usepackage{csvsimple-l3}   
\usepackage{xcolor}         
\usepackage{bm}
\usepackage{enumitem}

\newcommand{\vct}[1]{\bm{#1}} 
\newcommand{\mat}[1]{\mathbf{#1}} 

\usepackage{graphicx}
\usepackage{amsmath,amssymb}
\DeclareMathOperator*{\argmax}{arg\,max}

\graphicspath{{figures/}}

\title{Characterizing universal object representations\\across vision models}

\author{%
  Florian P.\ Mahner\thanks{Equal contribution.}\kern0.5em\thanks{Work mostly done while at the Max Planck Institute for Human Cognitive and Brain Sciences.}\\
  Machine Learning Group\\
  Technische Universit\"at Berlin\\
  BIFOLD\thanks{Berlin Institute for the Foundations of Learning and Data.}
  \And
  Johannes Roth\footnotemark[1]\\
  Vision and Computational Cognition Group\\
  MPI for Human Cognitive and Brain Sciences\\
  Justus-Liebig-University Giessen
  \AND
  Ka Chun Lam\\
  Machine Learning Core\\
  National Institute of Mental Health
  \And
  Michael F.\ Bonner\\
  Department of Cognitive Science\\
  Johns Hopkins University
  \AND
  Francisco Pereira\\
  Machine Learning Core\\
  National Institute of Mental Health
  \And
  Martin N.\ Hebart\\
  Vision and Computational Cognition Group\\
  MPI for Human Cognitive and Brain Sciences\\
  Justus-Liebig-University Giessen
}
\date{}

\begin{document}

\maketitle
\begin{abstract}
Deep neural networks trained with different architectures, objectives, and datasets have been reported to converge on similar visual representations. However, what remains unknown is which visual properties models actually converge on and which factors may underlie this convergence. To address this, we decompose the object similarity structure of 162 diverse vision models into a small set of non-negative dimensions. To determine universal versus model-specific dimensions, we then estimate how often each dimension reappears across models. In contrast to model-specific dimensions, universal dimensions are more interpretable and more strongly driven by conceptual image properties, indicating the relevance of interpretability and semantic content as implicit factors driving universality across models. Differences in architecture, objective function, training data, model size, and model performance do not explain the emergence of universal dimensions. However, models with more universal dimensions also better predict macaque IT activity and human similarity judgments, suggesting that universality reflects representations relevant to biological vision. These findings have important implications for understanding the emergent representations underlying deep neural network models and their alignment with biological vision.
\end{abstract}

\section{Introduction}

\begin{figure*}[tbp] 
    \centering
    \includegraphics[width=\textwidth]{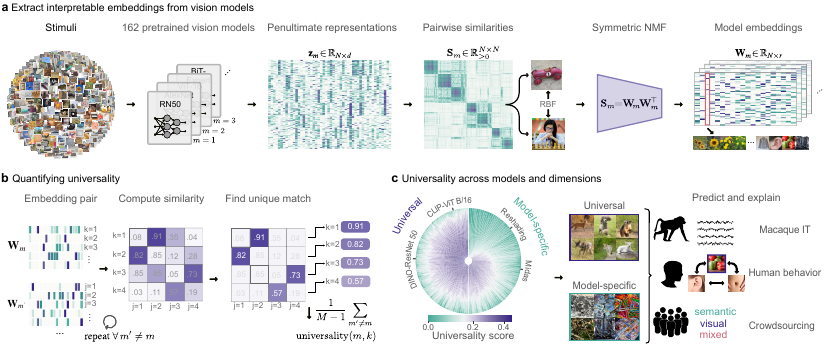}
    \caption{\textbf{Overview of the analysis pipeline and universality framework.} \textbf{(a)}~For each of 162 vision models, we extract penultimate-layer representations for object images, compute pairwise object similarity matrices, and apply symmetric nonnegative matrix factorization to obtain non-negative embeddings. \textbf{(b)}~We compute a universality score that quantifies how consistently each dimension of a model's embedding reappears across all other models. \textbf{(c)}~Left: distribution of per-dimension universality scores across all 162 models and 50 dimensions per model. Right: universal and model-specific dimensions are used to predict neural activity in macaque IT cortex and human similarity judgments, and to label the dimension content via online crowdsourcing.}
    \label{fig:overview}
\end{figure*}

Deep neural networks have become central to visual representation learning \citep{LeCun2015} and are increasingly used as computational models of the visual system \citep{Kanwisher2023, Doerig2023, Yamins2016}. Recent evidence suggests that deep neural networks trained on visual tasks converge on a common representational structure, despite substantial differences in architecture, objective function, and training data \citep{Huh2024, Guth2024}. This shared structure aligns more closely with neural responses in human visual cortex than model-specific structure \citep{Hosseini2024, Chen2025}, suggesting that these recurring representations may matter for biological vision, as well.

However, what properties these shared representations capture and what factors determine their emergence remain poorly understood. First, it is unclear whether universal representations reflect high-level conceptual properties such as object categories, or perceptual image properties such as edges and colors that have long been known to be shared across models \citep{Guth2024, Mahner2025}. Second, prior work has shown that training data and objective often matter more than architecture for alignment with brains and behavior \citep{Muttenthaler2023, Conwell2024}, but which of these factors drive universality across a large and diverse set of vision models, and how universal properties relate to biological vision, has remained open.

Addressing these questions requires us to overcome several challenges. First, we need to make representations across a wide range of different model architectures directly comparable. Second, we need to find dimensions underlying these representations that decompose an image representation at a level of granularity suitable for cross-model comparisons, since overly fine-grained representations (e.g.\ individual neurons) lack a canonical correspondence across models. We address these questions by decomposing the internal representations of 162 diverse vision models \citep{Conwell2024} for 22,248 object images from the THINGS database \citep{Hebart2019} into a small set of non-negative dimensions, following cognitive-science work showing that human similarity judgments for natural objects are captured by a similarly small, interpretable, non-negative representation \citep{Hebart2020}. We then quantify the universality of each dimension by measuring how consistently it recurs across models. Together, the loadings and universality score make it possible to characterize what properties make up universal and model-specific dimensions, which inductive biases are related to universality, and how universality aligns with biological vision. Our contributions are as follows:

\begin{itemize}
    \item We find that vision models, despite their differences, share a set of interpretable universal dimensions.
    \item We demonstrate that universal dimensions are more interpretable and more strongly explained by conceptual image properties than model-specific dimensions, which are less interpretable and more strongly driven by perceptual image properties.
    \item By isolating the role of architecture, objective function, and training data, we find that no single inductive bias accounts for universality, nor does model size or downstream performance.
    \item Instead, we show that universal dimensions are highly predictive of multi-unit activity in macaque inferior temporal cortex and of human behavioral similarity judgments, suggesting that universal dimensions share constraints with biological vision.
\end{itemize}

\section{Related work}
Our work relates to two broad research lines, one on representational alignment, and one on concept-based interpretability. Work on representational alignment has shown that vision models can differ substantially in what they learn: They rely more on texture than on shape relative to humans \citep{Geirhos2021, Hermann2020-01}, produce distinct intermediate representations across random seeds \citep{Li2015}, and differ in overall representational geometry \citep{Kornblith2019}. The Platonic Representation Hypothesis proposes that, despite differences in architecture, objective function, and training data, model representations become increasingly similar with scale and even across modalities \citep{Huh2024}, though recent work has questioned this: similarity measures can be inflated by model width and depth, leaving mainly local neighborhood structure shared \citep{Groger2026}, similar behavior can arise from very different representations \citep{Kumar2025}, and cross-modal alignment seems to emerge only when modalities share sufficient information \citep{Tjandrasuwita2025}. Other work has examined a related question within vision, asking whether diverse models share universal structure and whether these shared components are especially relevant for neural alignment \citep{Guth2024,Hosseini2024, Chen2025}. Across work relating models to neural responses and behavior, training data and objective function often matter more than architecture or scale once other factors are controlled \citep{Conwell2024, Wang2023, Prince2024, Chung2024, Muttenthaler2023, Muttenthaler2025}. These studies, however, mainly establish \textit{whether} models overlap, not \textit{what properties} the shared representations actually capture. Concept-based interpretability instead asks what models have learned in human-understandable terms \citep{Bau2017, Kim2018}, rather than merely visualizing the most predictive image regions \citep{Zeiler2014}. A useful distinction has been made between concept discovery, which extracts interpretable units from a representation, and concept importance estimation, which measures their contribution to model outputs \citep{Fel2023-01}. Early methods tested whether models represented fixed human-labeled concepts \citep{Bau2017, Kim2018}, while later work recovered concepts directly from activations \citep{Ghorbani2019, Zhang2021, Graziani2023, Vielhaben2023, Kowal2024, Fel2023, Klindt2023}. More recent work used sparse autoencoders to describe a model by a large set of directions in its activation space, where each direction corresponds to a concept the model has learned \citep{Cunningham2024, Bricken2023, Gao2025SAE, Fel2025, Bhalla2024}, and \citet{Thasarathan2025} extended this to multiple models by jointly training a single sparse autoencoder across a small set of pretrained models to recover a shared set of directions. These approaches operate in each model's activation space, so comparing many models requires aligning spaces and matching across thousands of features. We therefore describe each model by the similarity relationships its activations induce over a fixed set of objects, and decompose these similarities into a small number of non-negative dimensions defined as loadings over the same stimuli. Most closely related to our work, \citet{Hebart2020} showed that human similarity judgments can be explained by a sparse, non-negative embedding whose dimensions are reproducible, interpretable, and predictive of behavior. \citet{Mahner2025} derived sparse, non-negative embeddings for a small set of deep neural networks and found that their dimensions were dominated by visual rather than semantic properties. We extend this line of work to a large sample of vision models and introduce a measure of how consistently each dimension recurs across them.

\section{Methods}

\subsection{Notation}

Let $\mathcal{X} = \{x_1, \dots, x_N\}$ be a dataset of $N$ object images, and let $\mathcal{F} = \{f_1, \dots, f_M\}$ be a collection of $M$ vision models. For an image $x_i \in \mathcal{X}$, each model $f_m \in \mathcal{F}$ extracts a $d_m$-dimensional feature vector from its penultimate layer, denoted as $\mathbf{z}_m(x_i) \in \mathbb{R}^{d_m}$. Note that the dimensionality $d_m$ may vary across models. We used the penultimate layer because it is closest to the behavioral output and serves as the final stage of representation, integrating high-level conceptual and fine-grained perceptual information to produce an output. For each model $f_m$, we collected the representations of all images in $\mathcal{X}$ into a matrix $\mathbf{Z}_m \in \mathbb{R}^{N \times d_m}$, where the $i$-th row is given by $\mathbf{z}_m(x_i)^\top$.

\subsection{Dataset and models}
We used the THINGS image database \citep{Hebart2019}, which spans 1{,}854 basic-level object categories. For each object category, we selected 12 image exemplars, yielding a dataset $\mathcal{X}$ with a total of $N = 22{,}248$ images. We chose THINGS since it was designed to span a broad range of object concepts, is not part of common training sets, and has associated neural and behavioral data, making it well suited for evaluating learned representations against biological vision. Using THINGS, we extracted penultimate-layer representations from $M = 162$ vision models applied to the images in $\mathcal{X}$. Our model set $\mathcal{F}$ comprised those in \citet{Conwell2024} (excluding IPCL models), plus six OpenCLIP ViT-L/14 models \citep{Cherti2023}. The resulting set spans four architecture classes (99 convolutional, 51 transformer, 9 MLP-Mixer, and 3 hybrid), a range of objective functions (supervised classification, self-supervised learning, vision-language contrastive learning, and one untrained baseline), and diverse training datasets (including ImageNet, ImageNet-21k, YFCC15M, LAION, and Taskonomy). The full list of models is provided in Table~\ref{tab:models}. This diversity enables controlled comparisons that disentangle the effects of architecture, objective, and training data on our universality metric.

\subsection{Symmetric nonnegative factorization}\label{sec:snmf}
To make dimensions comparable across models with different feature bases and interpretable in terms of the images that define them, we characterize each model by the similarity structure it induces over a common image set and decompose that structure into a small set of nonnegative dimensions (Fig.~\ref{fig:overview}a).

\paragraph{Generate representational similarity matrix}
Models in $\mathcal{F}$ produce features in spaces of different dimensionalities and different bases, and several approaches exist for comparing such representations \citep{Kornblith2019, Williams2021}. We work at the level of similarity structure \citep{Kriegeskorte2008}, characterizing each model by the pairwise similarities it induces over the images in $\mathcal{X}$. This level of description is invariant to rotations of each model's feature basis and yields a common $N \times N$ matrix for every model in $\mathcal{F}$. Because similarity is determined by the pairwise distances between image representations, feature dimensions that vary little across $\mathcal{X}$ contribute little to these distances, so the similarity matrix emphasizes the structure along which the images actually differ. Similarity is defined over the shared image set, so any dimensions we extract from these matrices are automatically indexed by the same images across models and can be compared directly. For each model $f_m$, we computed a symmetric, entrywise nonnegative similarity matrix $\mat{S}_m \in \mathbb{R}_{\geq 0}^{N \times N}$ via a radial-basis function (RBF) kernel:

\begin{equation}
\label{eq:rbf}
[\mat{S}_m]_{i,j} = \exp\!\Bigl(-\frac{\lVert \vct{z}_m(x_i) - \vct{z}_m(x_j)\rVert^2}{2\sigma_m^2}\Bigr),
\end{equation}

where $\sigma_m$ is chosen per model to jointly maximize factorization stability and the explained variance of the low-rank reconstruction. Concretely, we searched over multipliers $\alpha$ of the median pairwise Euclidean distance $\tilde{d}_m$ \citep{Scholkopf1997}, setting $\sigma_m = \alpha^* \cdot \tilde{d}_m$, where $\alpha^*$ maximizes the harmonic mean of both criteria (Appendix~\ref{app:rbf}). We used the RBF kernel because it is a common choice for similarity measures \citep{Kornblith2019}, and it maps similarities to the interval $[0,1]$ and thereby supports interpretability for the subsequent nonnegative factorization step. Furthermore, the RBF kernel guarantees that $\mat{S}_m$ is positive semi-definite and nonnegative, both of which are required properties for symmetric nonnegative factorization (Eq.~\ref{eq:snmf}).

\paragraph{Decompose similarity matrix into interpretable image embeddings}
We decomposed each similarity matrix into $r$ non-negative dimensions using symmetric NMF~\citep{Lee2000}. For a chosen rank $r$, we estimated a nonnegative embedding $\mat{W}_m \in \mathbb{R}_{\geq 0}^{N \times r}$ of all images in $\mathcal{X}$ by solving:
\begin{equation}\label{eq:snmf}
  \min_{\mat{W}_m \geq 0}\; \tfrac{1}{2}\bigl\lVert \mat{S}_m - \mat{W}_m \mat{W}_m^{\!\top} \bigr\rVert_F^2.
\end{equation}

Each image is represented by a row of the matrix $\mat{W}_m$. Each column $\vct{w}_{m,k} \in \mathbb{R}_{\geq 0}^{N}$ defines a \emph{dimension}: its entries provide a nonnegative loading for each object image, and the numeric weight shows how strongly that image contributes to dimension $k$. The nonnegativity constraint promotes an interpretable parts-based factorization~\citep{Lee2000}. We optimized Eq.~\ref{eq:snmf} using block-successive upper-bound minimization~\citep{Shi2017}, following the authors' extension (Appendix~\ref{app:optimization}), independently for each model and for ranks $r \in \{10, 30, 50, 100, 200\}$. For each model, rank, and candidate bandwidth multiplier, we ran $B = 5$ random initializations. The bandwidth was selected by the harmonic mean of factorization stability and explained variance (Appendix~\ref{app:rbf}); within the selected bandwidth, we retained the most central seed (Appendix~\ref{app:optimization}). For all model comparisons, we fixed rank to $r=50$. In Appendix~\ref{app:rank_selection}, we show that our main results are insensitive to this choice.

\subsection{Universality}\label{sec:universality}
We quantified how consistently a dimension recurs across models (Fig.~\ref{fig:overview}b). Given a dimension~$k$ in model~$f_m$ and a dimension~$j$ in model~$f_{m'}$, we measured their agreement via squared cosine similarity, which is scale-invariant, bounded, and consistent with the Frobenius-norm objective of symmetric NMF:
\begin{equation}\label{eq:cos2}
  s(\mathbf{w}_{m,k},\, \mathbf{w}_{m',j})
  \;=\; \cos^2\!\bigl(\mathbf{w}_{m,k},\, \mathbf{w}_{m',j}\bigr)
  \;=\; \frac{(\mathbf{w}_{m,k}^\top \mathbf{w}_{m',j})^2}
             {\lVert \mathbf{w}_{m,k}\rVert^2\,\lVert \mathbf{w}_{m',j}\rVert^2}
  \;\in [0,1].
\end{equation}
Because symmetric NMF factors are identifiable only up to permutation, we established correspondences between models through a one-to-one assignment. Unlike a greedy best-match, which can allow multiple target dimensions to map onto the same source dimension, this formulation enforces a bijection across the full set of dimensions (Appendix~\ref{app:metric_details}). For each model pair $(m, m')$, we determined the optimal permutation $\pi^*_{m,m'}\!\in\!\arg\max_{\pi \in \mathcal{P}_r} \sum_{k=1}^{r} s(\mathbf{w}_{m,k},\, \mathbf{w}_{m',\pi(k)})$ via the Hungarian algorithm and defined the per-dimension universality as:

\begin{equation}\label{eq:universality}
  \operatorname{universality}(m, k)
  \;=\; \frac{1}{M-1}\sum_{m'\neq m}
        s\!\bigl(\mathbf{w}_{m,k},\;\mathbf{w}_{m',\pi^*(k)}\bigr).
\end{equation}

As all dimensions are nonnegative, any two of them will share some positive cosine similarity even if unrelated, and denser dimensions will have a higher baseline than sparser ones. This could distort the ranking of universality across dimensions. We corrected for this by calibrating against a permutation null that shuffles the row order of $\mathbf{W}_{m'}$, destroying stimulus correspondence while preserving column structure (Appendix~\ref{app:metric_details}). After calibration, values near~$1$ indicate dimensions that are consistently recovered across models, while values near~$0$ indicate dimensions with chance-level cross-model agreement. Universality is thus defined relative to the set of models $\mathcal{F}$ it is computed over, here 162 trained vision models (see Appendix~\ref{app:baselines} for untrained networks).
 
\section{Results}
\subsection{Universal dimensions are shared across vision models}

We first computed universality scores across 8{,}100 dimensions from 162 models. Scores ranged from near chance (0.0003) to 0.54 (Fig.~\ref{fig:overview}c), with the upper end reaching 64\% of the within-model stability ceiling, estimated from independent runs of the same model (median ceiling 0.84, Appendix~\ref{app:stability_ceiling}). Most models contained both highly universal and model-specific dimensions: pooling all 8{,}100 dimension scores, 80\% of models had at least one dimension in the top quartile and one in the bottom quartile. To confirm that these findings are not driven by the specific choice of images or models, we recomputed universality scores after (i)~changing the image set, (ii)~bootstrap-resampling the model set, and (iii)~excluding entire model families. As an alternative image set, we used ObjectNet~\citep{Barbu2019}, which depicts objects in cluttered real-world scenes with varied viewpoints and rotations, in contrast to the object-centered images in THINGS. Universality scores recomputed from ObjectNet embeddings largely preserved the model ranking ($\rho = 0.81$). Bootstrap resampling to 20\% of models ($n = 32$) yielded a highly consistent ranking ($\rho = 0.97$; Fig.~\ref{fig:stability}a), and excluding model families also left the ranking largely intact ($\rho = 0.83$; Fig.~\ref{fig:stability}b). These controls confirm that universality is a stable property of how models organize object representations (see Appendix~\ref{app:universality_validation} for details). In contrast, representations not shaped by training (random noise, image pixels, untrained networks) scored near the floor of the measure (Appendix~\ref{app:baselines}).

\subsection{Universal dimensions are biased towards interpretable, conceptual image properties}

\begin{figure*}[t]
    \centering
    \includegraphics[width=\textwidth]{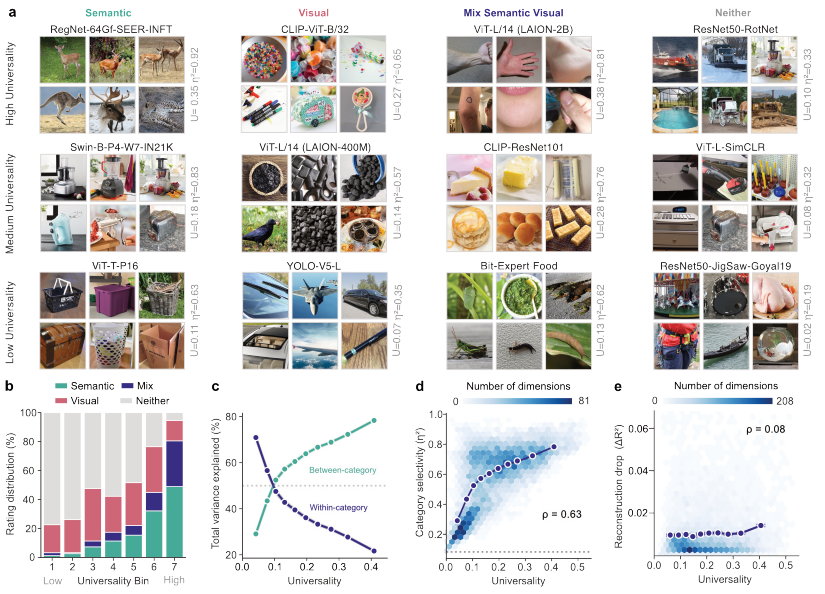}
    \caption{\textbf{Universal dimensions are more driven by conceptual object properties, and model-specific dimensions by visual properties or lack interpretable structure.} \textbf{(a)}~Representative dimensions for each label category as assigned by the majority of human raters (columns), at low, mid, and high universality (rows). Each grid shows a subset of the 64 top-weighted THINGS images that raters saw for that dimension. \textbf{(b)}~Proportion of dimensions assigned each human label (semantic, visual, mix semantic-visual, neither) across all universality bins. At low universality, 78\% of dimensions are rated as uninterpretable (\textit{neither}); at high universality, semantic and mixed dimensions together account for 80\%. \textbf{(c)}~Between- and within-category image loading variance across universality deciles, pooled over all 162 models. The most universal dimensions (top decile) are dominated by between-category variance ($\sim$78\%), while the most model-specific dimensions (bottom decile) are dominated by within-category variance ($\sim$71\%). \textbf{(d)}~Category consistency ($\eta^2$) vs.\ universality for all dimensions ($\rho = 0.63$). \textbf{(e)}~Reconstruction importance ($\Delta R^2$) vs.\ universality. The near-zero pooled correlation ($\rho = 0.01$ across all dimensions; within-model median $\rho = 0.08$) indicates that model-specific dimensions contribute as much to a model's similarity structure as universal ones.}
    \label{fig:category_selectivity}
\end{figure*}

We next asked what image properties distinguish universal from model-specific dimensions (Fig.~\ref{fig:category_selectivity}a). Some dimensions have high values for images sharing high-level object concepts such as animals, food, or household objects, while others group images by lower-level perceptual properties such as color or texture, and some show no clear pattern at all. To quantify the relationship between the dimension content and its universality, we asked crowd-sourced participants to label 1{,}059 representative dimensions, selected via clustering from the full set of 8{,}100. For each dimension, raters viewed its 64 top-weighted images and indicated whether they share an object or a semantic concept (\textit{semantic}), have a similar visual appearance without sharing an object or concept (\textit{visual}), both (\textit{both}), or whether they are uninterpretable (\textit{neither}; Appendix~\ref{app:dimension_ratings}). This is a commonly used distinction in research on object similarity \citep{Jozwik2016, Jozwik2017, Hebart2020, Deselaers2011} and on neural network representations \citep{Mahner2025}. Ratings were reliable, with median per-rater agreement with the majority label of 0.67 and median split-half agreement of 0.57. Binning the dimensions by their universality score revealed a sharp change in interpretability (Fig.~\ref{fig:category_selectivity}b). At low universality, 78\% of dimensions were labeled \textit{neither}, indicating that model-specific dimensions largely lack interpretable structure. At high universality this fraction collapsed to 4\%, indicating that universal dimensions were more interpretable to raters. Among the dimensions deemed interpretable, the content also varied with universality. The \textit{semantic} and \textit{both} labels together accounted for 80\% of dimensions at high universality, while the \textit{visual} label remained flat across the universality range. Thus, universal dimensions capture high-level object concepts, whereas the interpretable model-specific dimensions reflect low-level visual features. To test whether this reflects our choice of the penultimate layer, we repeated the analysis at three earlier network depths (Appendix~\ref{app:depth}). At every depth, more universal dimensions were more interpretable, and their semantic content increased with depth. Semantic features from word embeddings and WordNet, and descriptions generated by language models, confirmed the human labels and their relation to universality (Appendices~\ref{app:label_grounding} and~\ref{app:llm_descriptions}).

Moreover, if universal dimensions capture conceptual object structure, we would expect them to show stronger organization around basic object categories than model-specific ones. Because THINGS provides 12 exemplars for each of 1{,}854 categories, we can decompose each dimension's loading variance into between- and within-category components. High between-category variance indicates that the dimension groups images by object category, while high within-category variance indicates that it varies independently of category. We formalize this as category consistency $\eta^2$, the fraction of variance explained by category membership (see Appendix~\ref{app:category_consistency}):
\begin{equation}\label{eq:eta2}
  \eta^2 = \frac{\mathrm{SS}_{\mathrm{between}}}{\mathrm{SS}_{\mathrm{total}}}.
\end{equation}

A high $\eta^2$ indicates grouping by object category, whereas a low $\eta^2$ reflects within-category variation in visual features such as texture, viewpoint, or color. Category consistency ($\eta^2$) correlates strongly with universality across all 8,100 dimensions ($\rho = .63$; Fig.~\ref{fig:category_selectivity}d). To see how this relationship changes across the universality spectrum, we binned all dimensions into deciles and computed the fraction of variance that falls between versus within categories in each bin (Fig.~\ref{fig:category_selectivity}c). The most universal dimensions (top decile) are dominated by between-category variance ($\sim$78\%), while for the most model-specific dimensions (bottom decile), within-category variance dominates ($\sim$71\%). This confirms that category-level object structure, a key component of the semantic labels assigned by human raters, is strongly associated with universality.

Universal dimensions thus capture conceptual object structure and are more interpretable, but does this mean they play a larger role in a model's similarity structure? We computed each dimension's reconstruction importance, the drop in variance explained in the similarity matrix when that dimension is removed. Universality and reconstruction importance show near-zero correlation (within-model median $\rho = 0.08$; Fig.~\ref{fig:category_selectivity}e), indicating that universality reflects conceptual content but not contribution to similarity structure.

\subsection{Model inductive biases do not determine universality}
\label{sec:what_drives_universality}
 
\begin{figure*}[tbp]
    \centering
    \includegraphics[width=\textwidth]{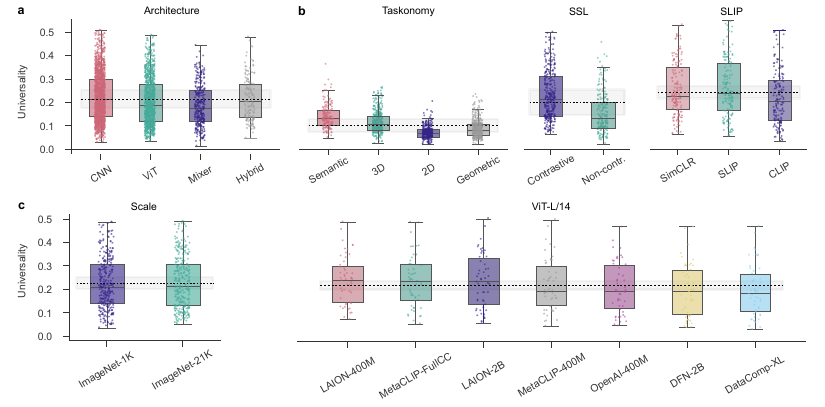}
    \caption{\textbf{Controlled comparisons of per-dimension universality.} Each panel varies one factor while holding the others approximately constant. Box plots show the distribution of per-dimension universality scores; individual dots are dimensions, colored by model. The dashed line and gray band indicate the grand mean $\pm$ 1 SD across all models in that group. \textbf{(a)}~Architecture: 34 CNNs, 21 transformers, 6 MLP-Mixers, and 3 hybrids, all trained on ImageNet-1K classification. \textbf{(b)}~Objective function: Taskonomy ResNet-50 encoders grouped by task cluster ($n = 23$, excluding two singleton clusters); contrastive vs.\ non-contrastive self-supervised ResNet-50 models ($n = 10$); SLIP ViT models trained with SimCLR, SLIP, or CLIP ($n = 9$). \textbf{(c)}~Training data. Left: ImageNet-1K vs.\ ImageNet-21K, matched by architecture ($n = 12$). Right: seven ViT-L/14 CLIP models trained on different web-scale datasets.}
    \label{fig:controlled}
\end{figure*}

If universality is a general property of learned visual representations, it should emerge across different architectures, objective functions, and training datasets rather than only under a specific design choice. To assess this, 
we first computed a single universality score per model, averaging across its 50 dimensions. This makes it possible to compare models that differ in one design choice while holding the others approximately constant. Model-level universality showed no reliable correlation with ImageNet top-1 accuracy ($\rho = -0.04$, $p = 0.72$; $n = 101$) or number of parameters ($r = 0.08$, $p = 0.31$; $n = 162$), ruling out model performance or size as trivial explanations (Fig.~\ref{fig:confounds}). Two-group contrasts were tested with Welch's t and multi-group contrasts with one-way ANOVA, with Bonferroni multiple comparison correction. We report Hedges' g for two-group contrasts and $\omega^2$ for multi-group contrasts with 95\% confidence intervals.

\paragraph{Architecture.}

We compared 34 CNNs, 21 transformers, 6 MLP-Mixers, and 3 hybrids, all trained on ImageNet-1K classification ($n = 64$). Architecture had no statistically significant effect on universality ($F(3, 60) = 2.54$, $p = 0.33$, $\omega^2 =0.07$ with 95\% CI [$-0.02$, $0.29$]; Fig.~\ref{fig:controlled}a), indicating that universality is not tied to a specific architecture class.

\paragraph{Objective function.}
We analyzed 25 ResNet-50 encoders from the Taskonomy benchmark~\citep{Zamir2019}, each trained on a different visual task with identical architecture and training data. These models showed low universality overall (cluster means ranging from 0.05 to 0.12, well below the full-set median of 0.22), reflecting the narrow distributions of their training tasks. Nevertheless, within this group, task cluster had a clear effect on the relative level of universality after excluding two singleton clusters ($F(3, 19) = 12.7$, $p < .001$, $\omega^2 = 0.60$; Fig.~\ref{fig:controlled}b, left). Semantic tasks (object/scene classification) produced the most universal dimensions, followed by 3D, Geometric, and 2D tasks. Across individual tasks, universality spanned a range of 0.125, from denoising at the bottom to object classification at the top. We next examined whether the choice of self-supervised learning objective affects universality. Among 10 self-supervised ResNet-50 models trained on ImageNet, contrastive objectives (SimCLR, MoCo, Barlow Twins, SwAV, DeepCluster) showed nominally higher universality than non-contrastive methods (RotNet, Jigsaw, ClusterFit), but the effect was not significant after correction (Welch $t = 3.21$, $p = .065$, Hedges' $g = 1.76$; Fig.~\ref{fig:controlled}b, center). Finally, to isolate the effect of language alignment, we compared ViT models from the SLIP family~\citep{Mu2022}, all trained on YFCC15M but with different objectives, SimCLR (purely visual), CLIP (vision-language), and SLIP (combined) ($n = 9$). Although CLIP models scored lower in universality than SimCLR and SLIP, the difference was not statistically significant ($F(2, 6) = 3.80$, $p = .43$, $\omega^2 = 0.38$; Fig.~\ref{fig:controlled}b, right).

\paragraph{Training data.}
We compared six models trained on ImageNet-1K with six trained on ImageNet-21K, matched by architecture family~\citep{Conwell2024}. Data scale had no effect on universality (Welch $t = -0.33$, $p > .99$, Hedges' $g = -0.18$; Fig.~\ref{fig:controlled}c, left). To further isolate the role of training data, we compared seven ViT-L/14 CLIP models that share the same architecture and objective but differ in training dataset (OpenAI WIT, LAION-400M, LAION-2B, DataComp-XL, DFN-2B, MetaCLIP-400M, MetaCLIP-FullCC). Universality scores within this group varied significantly less than expected from a random sample of seven models (bootstrap test on within-group standard deviation with null drawn from all 162 models, one-sided $p = .009$, 10{,}000 resamples; Fig.~\ref{fig:controlled}c, right). In a separate analysis of four IPCL models that differ only in training data \citep{Konkle2022}, the model trained on faces alone was least universal ($U = 0.056$) and models trained on diverse object images most universal ($U = 0.102$--$0.106$; Appendix~\ref{app:ipcl}). Together, neither dataset scale nor training-data choice substantially changed universality. Overall, although training objective may slightly modulate universality in specific model subsets, no single inductive bias accounts for universality. Universality thus emerges across diverse design choices, suggesting that it reflects a general property of learned visual representations.

\subsection{Universality predicts neural and human behavioral alignment}

\begin{figure*}[t]
    \centering
    \includegraphics[width=\textwidth]{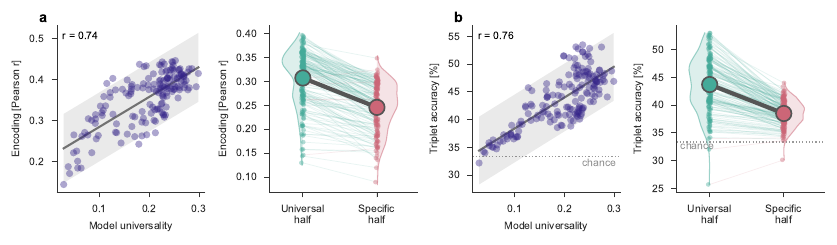}
    \caption{\textbf{Universality predicts neural and behavioral alignment.} \textbf{(a)}~Left: model universality vs IT encoding accuracy (mean across two macaques); Right IT encoding accuracy from the universal vs specific half of each model's dimensions. \textbf{(b)}~Left: model universality vs human triplet accuracy; Right: triplet accuracy from the universal vs specific half. Dotted lines are chance level (1/3).}
    \label{fig:encoding}
\end{figure*}

So far, the results have established that universal dimensions capture high-level conceptual object properties and that they emerge across diverse architectures, objective function, and training datasets rather than being driven by any single design choice. This leaves open why different models converge on these particular dimensions. A plausible explanation is that universal dimensions reflect aspects of the visual world itself rather than consequences of any particular model design. If so, universal dimensions should align with biological visual representations, since both are shaped by the same visual world, whereas model-specific dimensions reflect idiosyncrasies of particular models. We tested this using neural recordings from macaque inferior temporal (IT) cortex and human behavioral judgments, both collected on the THINGS image set.

\paragraph{Prediction of macaque IT neural recordings}
To quantify the alignment between model dimensions and neural activity, we predicted multi-unit activity in macaque IT cortex from each of the 162 models' dimensions using cross-validated ridge regression (see Appendix~\ref{app:neural_predictivity} for details). Models with more universal dimensions also predict IT activity more accurately ($r = 0.74$, $p < .001$; Fig.~\ref{fig:encoding}a). To test whether this is driven by universal dimensions specifically, we split each model's dimensions at the median universality score into a universal and a model-specific half. The universal half predicts IT activity significantly better than the model-specific half (paired $t$-test, $p < .001$).

\paragraph{Prediction of human behavior}
To quantify the alignment between model dimensions and behavior, we evaluated the 162 models against a dataset of odd-one-out judgments \citep{Hebart2020}, in which participants viewed triplets of object images and chose the image least similar to the other two. For each model, we predicted the human choice from pairwise cosine similarities between the embeddings $\mathbf{W}_m$ of the three images (see Appendix~\ref{app:behavioral_predictivity} for details). Models with more universal dimensions also predict human judgments more accurately ($r = 0.76$, $p < .001$; Fig.~\ref{fig:encoding}b), and the universal half of each model's dimensions yields significantly higher accuracy than the model-specific half ($0.42$ vs.\ $0.37$, chance $=0.33$, paired $t$-test, $p < .001$). Both relationships persist after controlling for each model's category consistency (Appendix~\ref{app:category_control}).

\paragraph{Human-aligned post-training}
If universal dimensions reflect structure that biological vision also captures, then making a model more human-like should make it more universal. We tested this with nine models post-trained toward human similarity judgments \citep{Muttenthaler2025}. Post-training increased universality in eight of nine models, along with IT and behavioral predictivity (Appendix~\ref{app:alignet}). Universality can thus be increased by aligning model representations with human behavior, suggesting a route to training more universal vision models.

\section{Discussion and conclusions}

We decomposed the representations of 162 vision models into interpretable dimensions using symmetric NMF and quantified how consistently each dimension reappears across models. Across this diverse set of models, we found that many dimensions appear repeatedly. These universal dimensions are also the ones that are most interpretable to human observers and most strongly tied to conceptual, between-category structure. In contrast, model-specific dimensions display higher within-category variation, are less interpretable, and relate more to lower-level visual properties. 
 
Our controlled comparisons show that no single inductive bias -- architecture, objective function, or training data -- accounts for universality. Taskonomy models show uniformly low universality relative to the broader model zoo, though, within this family, semantic tasks produce more universal dimensions than geometric or 2D tasks. Instead, universality is more closely aligned with biological vision than with any inductive bias. Universal dimensions predict both neural activity in macaque IT cortex and human behavioral similarity judgments more accurately than model-specific ones. The very same dimensions that align with brains and behavior are also the ones that human raters find interpretable and label as semantic. This alignment persists beyond what category membership explains, suggesting that brains, behavior, and models converge on the same conceptual structure rather than on a shared set of categories. Interpretability, conceptual content, and biological alignment thus co-occur in the structure that diverse models converge on. This suggests that universality is an emergent property of successful visual representation learning.

A growing body of work has provided evidence for convergence and shared structure across diverse models \citep{Huh2024, Chen2025, Hosseini2024}, while other work has questioned its extent \citep{Groger2026, Kumar2025, Tjandrasuwita2025}. Our results are consistent with the view that there may be universal dimensions of visual representation, with diverse models converging on conceptual object properties. However, this convergence is far from complete. The most universal dimensions do not have a perfect universality score, and model-specific dimensions, which capture lower-level visual properties, contribute equally to each model's representational geometry.

Prior work established that diverse models share representational structure that aligns with visual cortex \citep{Guth2024, Hosseini2024, Chen2025}, but did not explain what this structure contains. We find that it is conceptual, interpretable, and aligned with biological vision. Because universality is tied to neural and behavioral alignment and increases when models are made more human-like, it may also be a useful target for training vision models. Understanding why it emerges so consistently across very different architectures, objectives, and training regimes may help clarify the underlying principles shared by artificial and biological vision. More broadly, our approach generalizes beyond vision and can be applied to any set of models that process a common set of inputs, offering an approach for understanding what is shared and what is specific across learned representations.

\section{Limitations}\label{sec:limitations}

Several limitations and open questions remain. Our model set is broad but biased toward systems suited to tasks that humans find useful, and may not characterize the full space of possible visual representations. Similarly, our images contain everyday objects, and testing universality on complex scenes or on visual content with little human relevance is an important next step. The additive, non-negative decomposition also cannot recover features that encode concepts at multiple levels of abstraction \citep{Muttenthaler2025} or are stored in superposition \citep{Thasarathan2025, Fel2025}, and other methods may reveal additional structure in the model-specific components. Whether universal dimensions can be used to improve vision models, for instance, through representation alignment \citep{Sucholutsky2025} or universality-based regularization during training, is another open question worth pursuing.

\section*{Code and data availability}
Code is available at \url{https://github.com/florianmahner/universal-representations}. Non-negative embeddings for all 162 models on THINGS and ObjectNet, together with metadata, are available at \url{https://doi.org/10.5281/zenodo.23979573}.

\begin{ack}

The authors thank Aria Wang and Drew Linsley for helpful feedback on the manuscript and Klaus-Robert M\"uller for fruitful discussions that motivated further analyses.

\paragraph{Author contributions.}
F.P.M. and M.N.H. conceived the study. F.P.M., J.R., and F.P. designed the research. F.P.M. and K.L. developed the methodology and the mathematical framework. F.P.M. and J.R. conducted the experiments and performed the analyses. M.F.B., F.P., and M.N.H. provided critical feedback and contributed to the interpretation of results. F.P.M. and J.R. wrote the initial draft of the manuscript. All authors reviewed and edited the final manuscript.

\paragraph{\normalcolor Funding.}
F.P.M. was funded by the Deutsche Forschungsgemeinschaft (German Research Foundation, DFG, Project No. 564690550). F.P.M., J.R., and M.N.H. acknowledge support by a Max Planck Research Group grant of the Max Planck Society awarded to M.N.H\@. M.N.H. was supported by the ERC Starting Grant COREDIM (ERC-2021-STG-101039712), a LOEWE Start Professorship by the Hessian Ministry of Higher Education, Research, Science and the Arts, and the Deutsche Forschungsgemeinschaft (German Research Foundation, DFG) under Germany's Excellence Strategy (EXC 3066/1 ``The Adaptive Mind'', Project No. 533717223). This study used the high-performance computing resources of the Raven and Cobra Linux clusters at the Max Planck Computing \& Data Facility (MPCDF), Garching, Germany \href{https://www.mpcdf.mpg.de/services/supercomputing/}{https://www.mpcdf.mpg.de/services/supercomputing/}. K.L. and F.P. were supported in part by the Intramural Research Program of the National Institutes of Health (NIH) (ZIC MH002968). The contributions of the NIH author(s) were made as part of their official duties as NIH federal employees, are in compliance with agency policy requirements, and are considered Works of the United States Government. However, the findings and conclusions presented in this paper are those of the authors and do not necessarily reflect the views of the NIH or the U.S. Department of Health and Human Services. M.F.B. acknowledges support through a Research Fellowship for Experienced Researchers from the Humboldt Foundation. The funders had no role in study design, data collection and analysis, decision to publish or preparation of the manuscript.

\paragraph{Competing interests.} The authors declare no competing interests.

\end{ack}

\clearpage
{\small
\bibliographystyle{unsrtnat}
\bibliography{bibliography}
}

\clearpage

\appendix
\renewcommand{\thefigure}{S\arabic{figure}}
\renewcommand{\theHfigure}{supp.\arabic{figure}}
\setcounter{figure}{0}
\renewcommand{\thetable}{S\arabic{table}}
\renewcommand{\theHtable}{supp.\arabic{table}}
\setcounter{table}{0}

\label{app:method_details}

\section{Model overview}
\label{app:models}

Our set of 162 vision models spans variation in architecture, objective, scale, training set, and training task (Table~\ref{tab:models}). Universality scores are robust to substantial perturbations of the trained-model cohort considered here (Appendix~\ref{app:stability}).

{\footnotesize
\setlength{\tabcolsep}{3pt}
\begin{longtable}{p{3.4cm}llllp{2cm}}
\caption{\textbf{Overview of all 162 vision models.} Models are grouped by architecture class and sorted alphabetically within each group.}
\label{tab:models}\\
\toprule
Model & Class & Family & Objective & Task & Data\\
\midrule
\endfirsthead
\toprule
Model & Class & Family & Objective & Task & Data\\
\midrule
\endhead
\bottomrule
\endfoot
\begin{filecontents*}{models.csv}
Model,Class,Family,Objective,Task,Data
2D Keypoints,CNN,ResNet,Supervised,Taskonomy,Taskonomy
3D Keypoints,CNN,ResNet,Supervised,Taskonomy,Taskonomy
Abstraction,CNN,ResNet,Supervised,BiT-Expert,BiT Transfer
AlexNet,CNN,AlexNet,Supervised,Supervised,ImageNet-1K
Animal,CNN,ResNet,Supervised,BiT-Expert,BiT Transfer
Arthropod,CNN,ResNet,Supervised,BiT-Expert,BiT Transfer
Autoencoder,CNN,ResNet,Supervised,Taskonomy,Taskonomy
Bird,CNN,ResNet,Supervised,BiT-Expert,BiT Transfer
CLIP-ResNet101,CNN,ResNet,Vision-Language,CLIP,OpenAI-400M
CLIP-ResNet50,CNN,ResNet,Vision-Language,CLIP,OpenAI-400M
CSP-ResNet50,CNN,ResNet,Supervised,Supervised,ImageNet-1K
Camera Pose (Fixated),CNN,ResNet,Supervised,Taskonomy,Taskonomy
Camera Pose (Nonfixated),CNN,ResNet,Supervised,Taskonomy,Taskonomy
ConvNext-B,CNN,ConvNeXt,Supervised,Supervised,ImageNet-1K
ConvNext-B-IN21K,CNN,ConvNeXt,Supervised,Supervised,ImageNet-21K
ConvNext-L,CNN,ConvNeXt,Supervised,Supervised,ImageNet-1K
ConvNext-L-IN21K,CNN,ConvNeXt,Supervised,Supervised,ImageNet-21K
Curvatures,CNN,ResNet,Supervised,Taskonomy,Taskonomy
DINO-ResNet50,CNN,ResNet,Self-Supervised,Self-Supervised,ImageNet-1K
DLA34,CNN,DLA,Supervised,Supervised,ImageNet-1K
Denoising,CNN,ResNet,Supervised,Taskonomy,Taskonomy
DenseNet121,CNN,DenseNet,Supervised,Supervised,ImageNet-1K
ECA-NFNet-L0,CNN,NFNet,Supervised,Supervised,ImageNet-1K
EfficientNet-B1,CNN,EfficientNet,Supervised,Supervised,ImageNet-1K
EfficientNet-B3,CNN,EfficientNet,Supervised,Supervised,ImageNet-1K
Egomotion,CNN,ResNet,Supervised,Taskonomy,Taskonomy
Euclidean Depth,CNN,ResNet,Supervised,Taskonomy,Taskonomy
Faster-RCNN-ResNet50-FPN,CNN,R-CNN,Supervised,Detection,COCO
Flower,CNN,ResNet,Supervised,BiT-Expert,BiT Transfer
Food,CNN,ResNet,Supervised,BiT-Expert,BiT Transfer
GMLP-S16,CNN,gMLP,Supervised,Supervised,ImageNet-1K
GMixer-24,CNN,gMixer,Supervised,Supervised,ImageNet-1K
GhostNet100,CNN,GhostNet,Supervised,Supervised,ImageNet-1K
GoogleNet,CNN,Inception,Supervised,Supervised,ImageNet-1K
HardCoreNAS-A,CNN,HardCoreNAS,Supervised,Supervised,ImageNet-1K
HardCoreNAS-F,CNN,HardCoreNAS,Supervised,Supervised,ImageNet-1K
Inception-V3,CNN,Inception,Supervised,Supervised,ImageNet-1K
Inpainting,CNN,ResNet,Supervised,Taskonomy,Taskonomy
Instrument,CNN,ResNet,Supervised,BiT-Expert,BiT Transfer
Jigsaw,CNN,ResNet,Supervised,Taskonomy,Taskonomy
Keypoint-RCNN-ResNet50-FPN,CNN,R-CNN,Supervised,Segmentation,COCO
MNASNet1.0,CNN,MNASNet,Supervised,Supervised,ImageNet-1K
Mammal,CNN,ResNet,Supervised,BiT-Expert,BiT Transfer
Mask-RCNN-ResNet50-FPN,CNN,R-CNN,Supervised,Segmentation,COCO
MiDaS,CNN,MiDaS,Supervised,Depth,Depth Mix
MobileNet-V2,CNN,MobileNet,Supervised,Supervised,ImageNet-1K
MobileNet-V3-Large,CNN,MobileNet,Supervised,Supervised,ImageNet-1K
NF-Net-L0,CNN,NFNet,Supervised,Supervised,ImageNet-1K
NF-ResNet50,CNN,ResNet,Supervised,Supervised,ImageNet-1K
Object,CNN,ResNet,Supervised,BiT-Expert,BiT Transfer
Object Classification,CNN,ResNet,Supervised,Taskonomy,Taskonomy
Occlusion Edges,CNN,ResNet,Supervised,Taskonomy,Taskonomy
Point Matching,CNN,ResNet,Supervised,Taskonomy,Taskonomy
Random Weights,CNN,ResNet,Untrained,Untrained,Taskonomy
RegNet-128Gf-SEER,CNN,RegNet,Self-Supervised,SEER,Random-1B
RegNet-128Gf-SEER-INFT,CNN,RegNet,Self-Supervised,SEER,Random-1B
RegNet-32Gf-SEER,CNN,RegNet,Self-Supervised,SEER,Random-1B
RegNet-32Gf-SEER-INFT,CNN,RegNet,Self-Supervised,SEER,Random-1B
RegNet-64Gf-SEER,CNN,RegNet,Self-Supervised,SEER,Random-1B
RegNet-64Gf-SEER-INFT,CNN,RegNet,Self-Supervised,SEER,Random-1B
RegNetX-64,CNN,RegNet,Supervised,Supervised,ImageNet-1K
RegNetY-64,CNN,RegNet,Supervised,Supervised,ImageNet-1K
Relation,CNN,ResNet,Supervised,BiT-Expert,BiT Transfer
ResNet101,CNN,ResNet,Supervised,Supervised,ImageNet-1K
ResNet152,CNN,ResNet,Supervised,Supervised,ImageNet-1K
ResNet18,CNN,ResNet,Supervised,Supervised,ImageNet-1K
ResNet50,CNN,ResNet,Supervised,Supervised,ImageNet-1K
ResNet50-BarlowTwins,CNN,ResNet,Self-Supervised,Self-Supervised,ImageNet-1K
ResNet50-ClusterFit,CNN,ResNet,Self-Supervised,Self-Supervised,ImageNet-1K
ResNet50-DeepClusterV2,CNN,ResNet,Self-Supervised,Self-Supervised,ImageNet-1K
ResNet50-JigSaw-Goyal19,CNN,ResNet,Self-Supervised,Self-Supervised,ImageNet-1K
ResNet50-JigSaw-P100,CNN,ResNet,Self-Supervised,Self-Supervised,ImageNet-1K
ResNet50-MoCo-V2,CNN,ResNet,Self-Supervised,Self-Supervised,ImageNet-1K
ResNet50-PIRL,CNN,ResNet,Self-Supervised,Self-Supervised,ImageNet-1K
ResNet50-RotNet,CNN,ResNet,Self-Supervised,Self-Supervised,ImageNet-1K
ResNet50-SimCLR,CNN,ResNet,Self-Supervised,Self-Supervised,ImageNet-1K
ResNet50-SwAV,CNN,ResNet,Self-Supervised,Self-Supervised,ImageNet-1K
Reshading,CNN,ResNet,Supervised,Taskonomy,Taskonomy
RetinaNet-ResNet50-FPN,CNN,R-CNN,Supervised,Detection,COCO
Room Layout,CNN,ResNet,Supervised,Taskonomy,Taskonomy
SEResNext50-32x4D,CNN,SENet,Supervised,Supervised,ImageNet-1K
SKResNext50-32x4D,CNN,SENet,Supervised,Supervised,ImageNet-1K
Scene Classification,CNN,ResNet,Supervised,Taskonomy,Taskonomy
SemNASNet100,CNN,SemNASNet,Supervised,Supervised,ImageNet-1K
Semantic Segmentation,CNN,ResNet,Supervised,Taskonomy,Taskonomy
ShuffleNet-V2-x1.0,CNN,ShuffleNet,Supervised,Supervised,ImageNet-1K
SqueezeNet1.0,CNN,SqueezeNet,Supervised,Supervised,ImageNet-1K
Surface Normals,CNN,ResNet,Supervised,Taskonomy,Taskonomy
Texture Edges,CNN,ResNet,Supervised,Taskonomy,Taskonomy
Unsupervised 2.5D Segmentation,CNN,ResNet,Supervised,Taskonomy,Taskonomy
Unsupervised 2D Segmentation,CNN,ResNet,Supervised,Taskonomy,Taskonomy
VGG16,CNN,VGG,Supervised,Supervised,ImageNet-1K
Vanishing Point,CNN,ResNet,Supervised,Taskonomy,Taskonomy
Vehicle,CNN,ResNet,Supervised,BiT-Expert,BiT Transfer
Xception,CNN,Inception,Supervised,Supervised,ImageNet-1K
YOLO-V5-L,CNN,YOLO,Supervised,YOLO,COCO+VOC
YOLO-V5-M,CNN,YOLO,Supervised,YOLO,COCO+VOC
YOLO-V5-S,CNN,YOLO,Supervised,YOLO,COCO+VOC
Z-Buffer Depth,CNN,ResNet,Supervised,Taskonomy,Taskonomy
CLIP-ViT-B/16,ViT,ViT,Vision-Language,CLIP,OpenAI-400M
CLIP-ViT-B/32,ViT,ViT,Vision-Language,CLIP,OpenAI-400M
CLIP-ViT-L/14,ViT,ViT,Vision-Language,CLIP,OpenAI-400M
CoaT-Lite-Tiny,ViT,CoaT,Supervised,Supervised,ImageNet-1K
CrossViT-B,ViT,CrossViT,Supervised,Supervised,ImageNet-1K
DINO-ViT-B16,ViT,ViT,Self-Supervised,Self-Supervised,ImageNet-1K
DPT-Hybrid,ViT,DPT,Supervised,Depth,Depth Mix
DeiT-B-P16-224,ViT,DeiT,Supervised,Supervised,ImageNet-1K
JX-NesT-Tiny,ViT,NesT,Supervised,Supervised,ImageNet-1K
LeViT128,ViT,LeViT,Supervised,Supervised,ImageNet-1K
OpenCLIP-ViT-L/14 (DFN-2B),ViT,ViT,Vision-Language,CLIP,DFN-2B
OpenCLIP-ViT-L/14 (DataComp-XL),ViT,ViT,Vision-Language,CLIP,DataComp-XL
OpenCLIP-ViT-L/14 (LAION-2B),ViT,ViT,Vision-Language,CLIP,LAION-2B
OpenCLIP-ViT-L/14 (LAION-400M),ViT,ViT,Vision-Language,CLIP,LAION-400M
OpenCLIP-ViT-L/14 (MetaCLIP-400M),ViT,ViT,Vision-Language,CLIP,MetaCLIP-400M
OpenCLIP-ViT-L/14 (MetaCLIP-FullCC),ViT,ViT,Vision-Language,CLIP,MetaCLIP-FullCC
PiT-B-224,ViT,PiT,Supervised,Supervised,ImageNet-1K
PiT-T-224,ViT,PiT,Supervised,Supervised,ImageNet-1K
PoolFormer-S36,ViT,PoolFormer,Supervised,Supervised,ImageNet-1K
Swin-B-P4-W7,ViT,Swin,Supervised,Supervised,ImageNet-1K
Swin-B-P4-W7-IN21K,ViT,Swin,Supervised,Supervised,ImageNet-21K
Swin-L-P4-W7,ViT,Swin,Supervised,Supervised,ImageNet-1K
Swin-L-P4-W7-IN21K,ViT,Swin,Supervised,Supervised,ImageNet-21K
Swin-T-P4-W7,ViT,Swin,Supervised,Supervised,ImageNet-1K
TnT-P16-224,ViT,TnT,Supervised,Supervised,ImageNet-1K
ViT-B-CLIP,ViT,ViT,Vision-Language,SLIP,YFCC-15M
ViT-B-P16,ViT,ViT,Supervised,Supervised,ImageNet-1K
ViT-B-P16-IN21K,ViT,ViT,Supervised,Supervised,ImageNet-21K
ViT-B-P32,ViT,ViT,Supervised,Supervised,ImageNet-1K
ViT-B-P32-IN21K,ViT,ViT,Supervised,Supervised,ImageNet-21K
ViT-B-R50-S16-IN21K,ViT,ViT,Supervised,Supervised,ImageNet-21K
ViT-B-SLIP,ViT,ViT,Vision-Language,SLIP,YFCC-15M
ViT-B-SimCLR,ViT,ViT,Self-Supervised,SLIP,YFCC-15M
ViT-L-CLIP,ViT,ViT,Vision-Language,SLIP,YFCC-15M
ViT-L-CLIP-CC12M,ViT,ViT,Vision-Language,SLIP,YFCC-15M
ViT-L-P16,ViT,ViT,Supervised,Supervised,ImageNet-1K
ViT-L-P16-IN21K,ViT,ViT,Supervised,Supervised,ImageNet-21K
ViT-L-SLIP,ViT,ViT,Vision-Language,SLIP,YFCC-15M
ViT-L-SLIP-CC12M,ViT,ViT,Vision-Language,SLIP,YFCC-15M
ViT-L-SimCLR,ViT,ViT,Self-Supervised,SLIP,YFCC-15M
ViT-S-CLIP,ViT,ViT,Vision-Language,SLIP,YFCC-15M
ViT-S-P16,ViT,ViT,Supervised,Supervised,ImageNet-1K
ViT-S-P16-IN21K,ViT,ViT,Supervised,Supervised,ImageNet-21K
ViT-S-P32,ViT,ViT,Supervised,Supervised,ImageNet-1K
ViT-S-P32-IN21K,ViT,ViT,Supervised,Supervised,ImageNet-21K
ViT-S-SLIP,ViT,ViT,Vision-Language,SLIP,YFCC-15M
ViT-S-SimCLR,ViT,ViT,Self-Supervised,SLIP,YFCC-15M
ViT-T-P16,ViT,ViT,Supervised,Supervised,ImageNet-1K
Visformer,ViT,Visformer,Supervised,Supervised,ImageNet-1K
XCIT-N-12-P16,ViT,XCiT,Supervised,Supervised,ImageNet-1K
XCIT-N-12-P8,ViT,XCiT,Supervised,Supervised,ImageNet-1K
MLP-Mixer-B16,Mixer,MLP-Mixer,Supervised,Supervised,ImageNet-1K
MLP-Mixer-B16-IN21K,Mixer,MLP-Mixer,Supervised,Supervised,ImageNet-21K
MLP-Mixer-L16,Mixer,MLP-Mixer,Supervised,Supervised,ImageNet-1K
MLP-Mixer-L16-IN21K,Mixer,MLP-Mixer,Supervised,Supervised,ImageNet-21K
ResMLP-12,Mixer,MLP-Mixer,Supervised,Supervised,ImageNet-1K
ResMLP-24,Mixer,MLP-Mixer,Supervised,Supervised,ImageNet-1K
ResMLP-36,Mixer,MLP-Mixer,Supervised,Supervised,ImageNet-1K
ResMLP-Big-24,Mixer,MLP-Mixer,Supervised,Supervised,ImageNet-1K
ResMLP-Big-24-IN21K,Mixer,MLP-Mixer,Supervised,Supervised,ImageNet-21K
ConViT-B,Hybrid,ConViT,Supervised,Supervised,ImageNet-1K
ConViT-T,Hybrid,ConViT,Supervised,Supervised,ImageNet-1K
ConvMixer-768-32,Hybrid,ConvMixer,Supervised,Supervised,ImageNet-1K
\end{filecontents*}
\csvreader[
  head to column names,
  late after line=\\,
]{models.csv}{}{\Model & \Class & \Family & \Objective & \Task & \Data}
\end{longtable}
}
\FloatBarrier

\section{Embedding generation}

\FloatBarrier
\subsection{RBF kernel and bandwidth selection}
\label{app:rbf}

We define the symmetric similarity matrix $\mat{S}_m\in\mathbb{R}^{n\times n}_{\ge 0}$ via a radial basis function (RBF) kernel,
\begin{equation}
\label{eq:rbf_appendix}
[\mat{S}_m]_{ij}
=\exp\!\Bigl(-\frac{\lVert \vct{z}_{m,i}-\vct{z}_{m,j}\rVert^2}{2\sigma_m^2}\Bigr),
\end{equation}
where the RBF kernel guarantees that $\mat{S}_m$ is positive semi-definite and nonnegative, both properties required by the symmetric nonnegative factorization (Eq.~\ref{eq:snmf}).

\begin{figure}[!htbp]
  \centering
  \includegraphics[width=\linewidth]{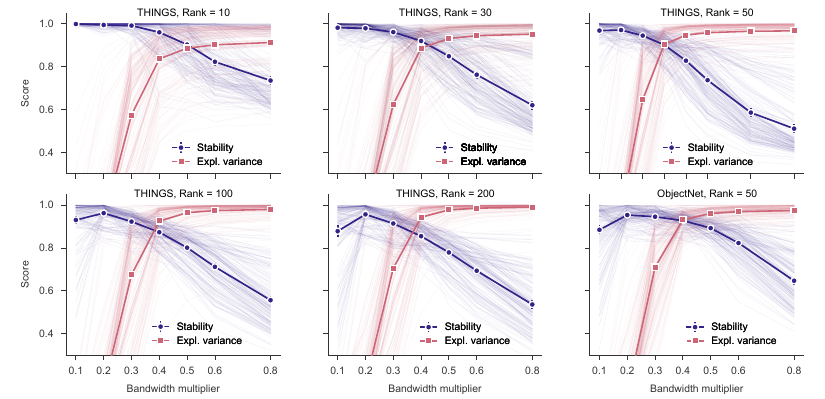}
  \caption{\textbf{Bandwidth selection.} Factorization stability and explained variance as a function of the RBF bandwidth multiplier $\alpha$, for THINGS ranks 10, 30, 50, 100, and 200, and ObjectNet rank 50. Thin lines show individual models ($n = 162$); thick lines show means $\pm$ 95\% CI. Dashed vertical lines indicate the optimal $\alpha^*$ (maximizing the harmonic mean of
both criteria). The optimum is consistently near $\alpha = 0.4$--$0.5$ across ranks.}
  \label{fig:sigma_selection}
\end{figure}

The bandwidth $\sigma_m$ controls the emphasis on local versus global similarity structure. Rather than fixing it to the median pairwise distance, we set $\sigma_m = \alpha^* \cdot \tilde{d}_m$, where $\tilde{d}_m$ is the median pairwise Euclidean distance within model $m$ \citep{Scholkopf1997} and $\alpha^*$ is chosen per model to jointly
maximize factorization stability and explained variance. Specifically, we search over a grid of multipliers $\alpha \in \{0.1, 0.2, 0.3, 0.4, 0.5, 0.6, 0.8, 1.0\}$, run $B = 5$ random initializations per multiplier, and select the $\alpha^*$ that maximizes the harmonic mean of factorization stability (mean pairwise correlation of aligned solutions across seeds) and explained variance of the low-rank reconstruction, thus penalizing imbalanced solutions in which one criterion is optimal because another is sacrificed. Fig.~\ref{fig:sigma_selection} shows that factorization stability decreases monotonically with $\alpha$ while explained variance increases, producing a consistent optimum near $\alpha^* = 0.4$--$0.5$ across all ranks.

Note that we have
\begin{equation}
     \frac{\partial[\mathbf{S}_m]_{ij}}{ \partial \alpha } = \frac{\partial}{\partial \alpha} \exp\left(- \frac{\Vert \vct{z}_{m,i} - \vct{z}_{m,j}\Vert^2}{2\alpha^2 \tilde{d}_m^2}\right) = [\mathbf{S}_m]_{ij} \cdot \frac{ \Vert \vct{z}_{m,i} - \vct{z}_{m,j}\Vert^2 }{\alpha^3 \tilde{d}_m^2} > 0
\end{equation}
which implies when $\alpha$ increases, every off-diagonal similarity increases monotonically toward 1. More specifically, we can rewrite $\mathbf{S}_m$ with Taylor expansion around large $\alpha$ as
\begin{equation}
    \mathbf{S}_m = \mathbf{1}\mathbf{1}^{\top} - \frac{1}{2\alpha^2 \tilde{d}_m^2} \mathbf{D} + O(\alpha^{-4})
\end{equation}
where $[\mathbf{D}]_{ij} = \Vert \vct{z}_{m,i} - \vct{z}_{m,j} \Vert^2$ is the squared pairwise Euclidean distance matrix. When $\alpha$ increases, less variance remains in higher-order directions, and $\mathbf{S}_m$ converges towards the leading term, which is rank 1. Consequently, for any fixed factorization rank $r$, the rank-$r$ symmetric NMF can capture a larger fraction of similarity structure, leading to higher explained variance.
At the same time, $\mathbf{S}_m$ contains less sharply differentiated structure as many pairs become similarly close. The matrix is dominated by broad global similarity. The optimization landscape for symmetric NMF becomes flatter in directions corresponding to splitting or merging of factors. The $WW^{\top}$ decomposition is then weakly constrained and therefore stability across runs drops.

\FloatBarrier
\subsection{Optimization}
\label{app:optimization}

We optimize Eq.~\ref{eq:snmf} via block successive upper-bound minimization \citep{Shi2017}. For each model $m$, rank $r$, and candidate bandwidth multiplier, we run $B=5$ random initializations. For each multiplier, we align solutions across initializations via the Hungarian algorithm and compute factorization stability as the mean matched correlation across seed pairs. After selecting the bandwidth multiplier by the harmonic mean of factorization stability and explained variance (Appendix~\ref{app:rbf}), we retain the most central seed, defined as the solution with the highest average matched correlation to all other seeds.

\FloatBarrier
\subsection{Rank selection}
\label{app:rank_selection}

\begin{figure}[!htbp]
    \centering
    \includegraphics[width=\textwidth]{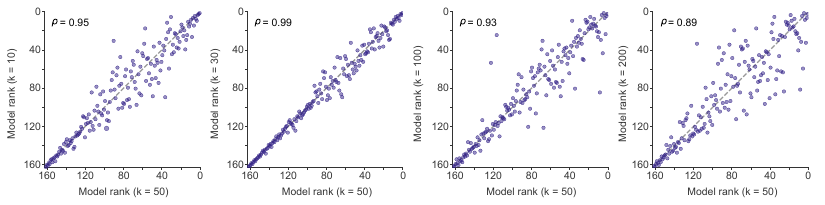}
    \caption{\textbf{Rank stability.} Model-level universality rankings at rank 50 compared against ranks 10, 30, 100, and 200. Each dot is one model; Spearman correlations are shown per panel.}
    \label{fig:rank_selection}
\end{figure}

We report results at rank $r = 50$ throughout the main text but repeat all analyses at $r \in \{10, 30, 100, 200\}$. Model-level universality rankings are highly stable across ranks (Fig.~\ref{fig:rank_selection}), with Spearman correlations between rank-50 and all other ranks at least $\rho = 0.93$ except for $r = 200$. This confirms that our conclusions do not depend on the particular factorization rank.

\FloatBarrier

\section{Metric validation}

\FloatBarrier
\subsection{Universality metric details}
\label{app:metric_details}

Our universality metric is used to test factor identity: whether an individual dimension recovered from one model recurs as an individual dimension in another model. This means that matching should be one-to-one, and scores should be calibrated against chance stimulus-level correspondence.

\begin{figure}[!htbp]
    \centering
    \includegraphics[width=\linewidth]{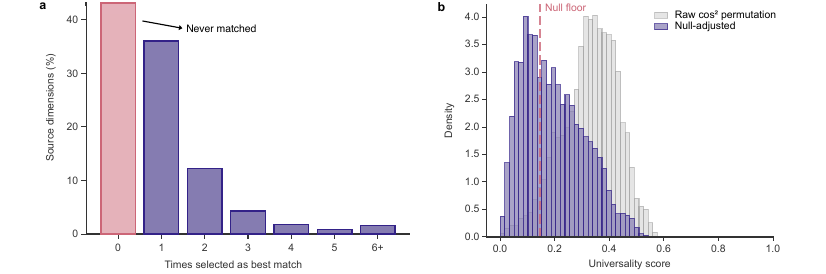}\\[2pt]
    \includegraphics[width=0.5\linewidth]{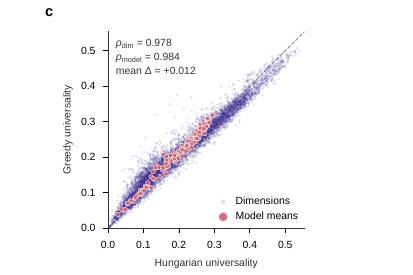}
    \caption{\textbf{Universality metric calibration and comparison with greedy matching.} \textbf{(a)}~Distribution of how often each source dimension is selected as a target dimension's best match across all model pairs at rank 50. Under greedy matching, 43\% of source dimensions are never selected, while others are claimed by multiple target dimensions. The Hungarian algorithm enforces exactly one match per dimension. \textbf{(b)}~Distribution of per-dimension universality scores before (gray) and after (blue) null correction. Raw $\cos^2$-permutation scores exhibit a positive floor from chance alignment (dashed line); null adjustment removes this floor and rescales scores to $[0,1]$. \textbf{(c)}~Greedy vs.\ Hungarian universality for all dimensions (blue) and model means (rose). The dashed line denotes identity.}
    \label{fig:universality_metric}
\end{figure}

\paragraph{\normalcolor Why not greedy matching?}
A natural alternative is independent best-match (greedy) matching, in which each target dimension selects its most similar source dimension without removing previously selected ones. This permits many-to-one collisions. Across all 26{,}082 model pairs at rank 50, 43\% of source dimensions are never selected as any target dimension's best match, while others are selected repeatedly (Fig.~\ref{fig:universality_metric}, a). Recomputing universality with greedy instead of Hungarian matching, the two agree closely ($\rho = 0.978$ across dimensions, $0.984$ across models; Fig.~\ref{fig:universality_metric}, c), so none of our conclusions depend on this choice. Greedy matching inflates scores, but nearly uniformly across models ($0.012$ on average, 95\% CI [0.010, 0.013]) and independently of a model's Hungarian score ($r = 0.001$), so no model benefits disproportionately. We use Hungarian matching because it is more conservative, matching every dimension exactly once.

\paragraph{Null calibration.}
At moderate ranks, even unrelated dimensions can achieve non-trivial $\cos^2$ scores by chance. To remove this floor, we construct a permutation null for each target model $m$: we randomly permute the rows of source embeddings $\mathbf{W}_{m'}$, destroying stimulus correspondence while preserving column structure, recompute the Hungarian assignment and $\cos^2$-permutation scores on each shuffled dataset, and take the 95th percentile across $B = 1{,}000$ permutations as a per-dimension threshold $a_{m,k}$. The null-adjusted score is
\begin{equation}
    s_{\mathrm{adj}}(m,k;\,m') = \frac{s_{\pi^*}(m,k;\,m') - a_{m,k}}{1 - a_{m,k}},
    \quad \text{clipped to } [0,1],
\end{equation}
and the final universality score averages these adjusted values across all remaining models. This calibration shifts the score distribution leftward and removes the positive floor present in the raw $\cos^2$-permutation scores (Fig.~\ref{fig:universality_metric}, b), ensuring that only dimensions with stimulus-specific correspondence contribute to universality.

\paragraph{Factor identity versus shared subspaces.}
The permutation-based universality score tests whether each factor in model $m$ has a unique counterpart in model $m'$. Alternatively, one can ask whether a factor lies in the nonnegative cone of another model's factors, by projecting $\mathbf{w}_{m,k}$ onto the cone $\mathcal{C}(\mathbf{W}_{m'}) = \{\mathbf{W}_{m'}\mathbf{a} : \mathbf{a} \geq 0\}$ via nonnegative least squares and measuring how much variance is captured. This cone projection variant captures shared representational subspaces rather than strict factor identity, including cases where one model splits or merges concepts across dimensions. However, while this can detect broader shared structure, it also mixes level of description: the source is an individual factor, while the target is a model-level subspace (or vice-versa). Consequently, a high cone score is ambiguous, as it may reflect a true one-to-one counterpart, a split/merge correspondence across multiple factors, or only a diffuse approximation by several partially related factors. We use permutation matching as our primary metric because it directly tests factor identity, that is, whether the same dimension recurs across models.

\FloatBarrier
\subsection{Within-model stability ceiling}
\label{app:stability_ceiling}

To interpret universality on an absolute scale, we compute the within-model stability ceiling, defined as the typical agreement between independent NMF factorizations of the same model. Using the $B = 5$ seeds fit per model (Appendix~\ref{app:optimization}), we compute the $\cos^2$-permutation score between each of the $\binom{5}{2} = 10$ seed pairs, average across pairs per dimension, and apply the same null calibration as for universality (Appendix~\ref{app:metric_details}).

The null-adjusted within-model stability has median 0.84 (IQR [0.75, 0.90], range [0.23, 0.995]) across all 8{,}100 dimensions. This is an empirical upper bound for cross-model agreement under $\cos^2$-perm, since a dimension cannot be more consistent across different models than it is across independent fits of a single model. Note that this ceiling varies only the NMF initialization while holding the underlying model fixed. A stricter ceiling that also varied model training, for example across independent training reruns of the same architecture, would likely be lower, making our estimate a liberal upper bound on cross-model agreement.

\FloatBarrier
\subsection{Universality metric validation}
\label{app:universality_validation}

\begin{figure}[!htbp]
    \centering
    \includegraphics[width=\linewidth]{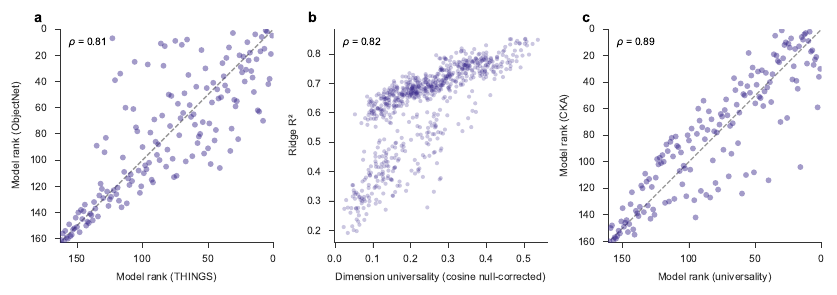}
    \caption{\textbf{Metric validation.} \textbf{(a)}~Model-level universality rankings computed from THINGS vs.\ ObjectNet features ($\rho = 0.81$). \textbf{(b)}~Per-dimension universality (null-corrected cosine) vs.\ cross-validated ridge $R^2$ ($\rho = 0.82$), confirming that our metric agrees with a more expensive regression-based proxy. \textbf{(c)}~Model-level universality rank vs.\ CKA rank ($\rho = 0.89$), establishing convergent validity with a standard representational similarity measure.}
    \label{fig:metric_validation}
\end{figure}

We summarize universality at the model level as $U_m=\frac{1}{r}\sum_{k=1}^r u_{m,k}$ and validate this metric in three ways (Fig.~\ref{fig:metric_validation}). First, we test whether universality generalizes to a different image set by recomputing scores from ObjectNet~\citep{Barbu2019}, which depicts objects in cluttered real-world scenes; model rankings are largely preserved. Second, we show that the cos$^2$-permutation metric agrees closely with a cross-validated ridge regression variant at the per-dimension level. Third, we confirm that universality correlates strongly with centered kernel alignment (CKA), establishing convergent validity across different representational similarity measures.

\FloatBarrier
\subsection{Universality vs.\ model accuracy and size}
\label{app:confounds}

\begin{figure}[!htbp]
    \centering
    \includegraphics[width=\linewidth]{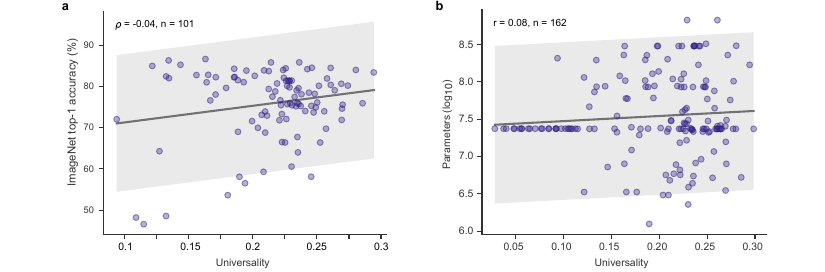}
    \caption{\textbf{Universality is not explained by model accuracy or size.} \textbf{(a)}~Universality vs.\ ImageNet top-1 accuracy ($n = 101$; Spearman $\rho = -0.04$, $p = 0.72$). \textbf{(b)}~Universality vs.\ number of parameters on a log scale ($n = 162$; Pearson $r = 0.08$, $p = 0.31$). Neither variable shows a reliable association with universality.}
    \label{fig:confounds}
\end{figure}

Universality could in principle reflect model quality rather than shared representational structure. To rule this out, we correlate per-model universality with ImageNet top-1 accuracy (for the 101 models with reported accuracy) and with the total number of parameters (all 162 models). Neither shows a reliable association (Fig.~\ref{fig:confounds}). The Pearson correlation with accuracy is weak and non-significant ($r = 0.19$, $p = 0.052$), and the rank correlation is essentially zero ($\rho = -0.04$, $p = 0.72$), indicating that the marginal Pearson value is driven by a small number of outliers. The Pearson correlation with parameter count is similarly negligible ($r = 0.08$, $p = 0.31$). Models that are more accurate or larger are not systematically more universal.

\FloatBarrier
\subsection{Model-set stability}
\label{app:stability}

We test whether the universality metric is overly dependent on the specific model set over which it is computed.

\paragraph{Subsampling stability.}
We subsample the model set to 20\% of models ($n = 32$, 1{,}000 iterations) and recompute universality scores. Model rankings based on overall universality are stable across iterations, with a median Spearman correlation between the full-set and subsampled rankings of $\rho = 0.97$ (95\% range: $[0.84, 0.99]$).

\paragraph{Leave-family-out stability.}
We also test whether the universality of dimensions in a given model depends on other models from the same architectural family being present in the set. Recomputing universality using only models from other architecture classes yields scores that are highly correlated with those of the full set ($\rho = 0.83$).

\begin{figure}[!htbp]
    \centering
    \includegraphics[width=\linewidth]{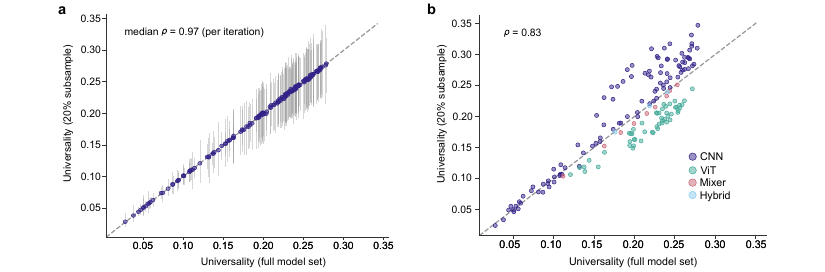}
    \caption{\textbf{Model-set stability.} \textbf{(a)}~Bootstrap stability: universality from the full model set ($M=162$) vs the mean across 1{,}000 bootstrap resamples (20\% of models, $n = 32$). \textbf{(b)}~Leave-family-out stability: universality from the full set vs recomputed after excluding all models from the same architecture family.}
    \label{fig:stability}
\end{figure}
\FloatBarrier

\section{Control baselines}
\label{app:baselines}

\begin{figure}[!htbp]
    \centering
    \includegraphics[width=\linewidth]{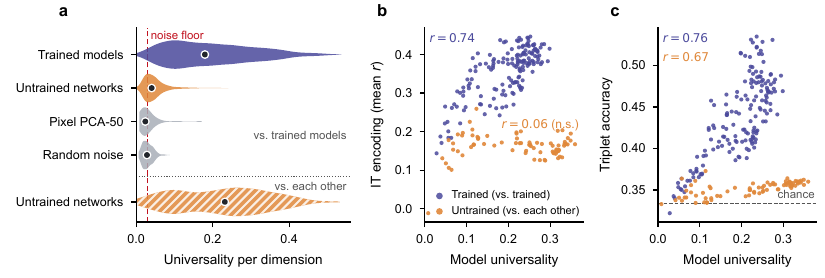}
    \caption{\textbf{Control baselines.} \textbf{(a)}~Per-dimension universality of trained models, untrained networks, pixel principal components, and Gaussian noise, scored against the 162 trained models; bottom row: untrained networks scored against each other. \textbf{(b, c)}~IT encoding (b) and triplet accuracy (c) vs.\ model universality for trained and untrained networks. The untrained outlier near zero is ConvMixer-768/32.}
    \label{fig:baselines}
\end{figure}

Here, we assess universality in random noise, image pixels, and untrained networks, providing a stringent test of the extent to which the shared structure found in trained models depends on learning. All three baselines were computed with the same pipeline and scored against the 162 trained models (Fig.~\ref{fig:baselines}a).

\paragraph{Random noise.}
First, we simulated representations from five independent draws of Gaussian noise, so that neither images nor models were involved. The resulting universality score was $U = 0.030$ ($\mathrm{SD} = 0.014$). This noise floor remained slightly above zero by construction. Each raw score was compared against the 95th percentile of a permutation null, and scores below this threshold were set to zero (Appendix~\ref{app:metric_details}). Thus, about 5\% of comparisons can exceed the threshold by chance. Because scores below the threshold are set to zero, these chance matches leave a small positive average even for random noise.

\paragraph{Pixels.}
Second, we considered pixel information from the images, without a model. If universality merely recovered the strongest dimensions of the input, a representation built from the input alone should be about as universal as a trained model. We computed the first 50 principal components of the raw pixels, derived a similarity matrix between images in that space, and computed an embedding from this matrix. Universality was again near the noise floor ($U = 0.031$, $\mathrm{SD} = 0.027$), compared with trained models ($U = 0.196$, $\mathrm{SD} = 0.112$; Fig.~\ref{fig:baselines}a).

\paragraph{Untrained networks.}
Third, we tested the degree to which universality arises through model training. We extracted penultimate-layer representations from randomly initialized versions of all 79 distinct architectures and repeated our analyses. Scored against the trained models, untrained dimensions had a mean universality of $U = 0.047$ ($\mathrm{SD} = 0.029$), slightly above the noise floor of $U = 0.030$. Among themselves, however, untrained networks recovered shared dimensions far above chance ($U = 0.223$, $\mathrm{SD} = 0.117$; Fig.~\ref{fig:baselines}a). This suggests that processing visual input through different untrained models can induce shared structure under the same image statistics, but that this structure differs strongly from the dimensions of learned representations.

We then asked how well the universal dimensions shared among untrained models explain macaque IT activity and human similarity judgments. Universality computed among untrained models was essentially unrelated to IT encoding performance ($r = 0.06$, $p = 0.61$), whereas universality among trained models strongly predicted IT encoding ($r = 0.74$; Fig.~\ref{fig:baselines}b). Untrained models also predicted IT activity worse overall than trained models (mean encoding correlations of $r = 0.162$ and $r = 0.355$, respectively). They predicted human similarity judgments slightly above chance, but far below trained models (mean accuracies of $0.352$ and $0.438$, respectively; chance level of $0.333$). Both trained--untrained differences were significant (Welch tests, both $p < .001$). Nevertheless, universality among untrained models predicted odd-one-out accuracy ($r = 0.67$, $p < .001$; Fig.~\ref{fig:baselines}c). Architectural constraints and image statistics may therefore contribute to shared dimensions that predict behavior, but we found no comparable relationship between universality and IT encoding in untrained models.

\FloatBarrier

\section{Human dimension rating experiment}
\label{app:dimension_ratings}

\paragraph{Goal.}
To characterize what each NMF dimension encodes from a human perspective, we collected crowd-sourced labels indicating whether each dimension reflects semantic content, visual properties, both, or neither.

\paragraph{Dimension selection.}
We applied hierarchical agglomerative clustering to all $162 \times 50 = 8{,}100$ dimensions using the precomputed linkage matrix (correlation distance). Re-cutting the dendrogram at a threshold of $0.75$ yielded 1{,}059 clusters. For each cluster, we selected the single representative dimension whose weight vector $\vct{w}$ had highest correlation with the cluster centroid (mean weight vector), reducing redundancy while preserving the diversity of dimension types.

\paragraph{Stimuli.}
For each selected dimension, we constructed an image grid showing the 64 highest-loading images, with at most one image per THINGS category (to prevent any single category from dominating). Images were sorted in descending order of loading weight.

\paragraph{Task.}
Participants were shown image grids one at a time and asked to categorize each dimension by selecting one of four mutually exclusive labels:
\begin{itemize}[noitemsep,topsep=2pt]
    \item \textbf{Semantic} --- the images share an identifiable object category or concept (e.g.\ all dogs, all vehicles).
    \item \textbf{Visual} --- the images share a low-level perceptual property such as color, texture, or shape, without a clear shared concept.
    \item \textbf{Both} --- the images share both a conceptual and a perceptual property.
    \item \textbf{Neither} --- no clear shared property is apparent.
\end{itemize}
Participants also rated the difficulty of labeling each dimension on a 1--7 scale ($1 = $ very easy, $7 = $ very difficult).

\begin{figure}[!htbp]
    \centering
    \includegraphics[width=\linewidth]{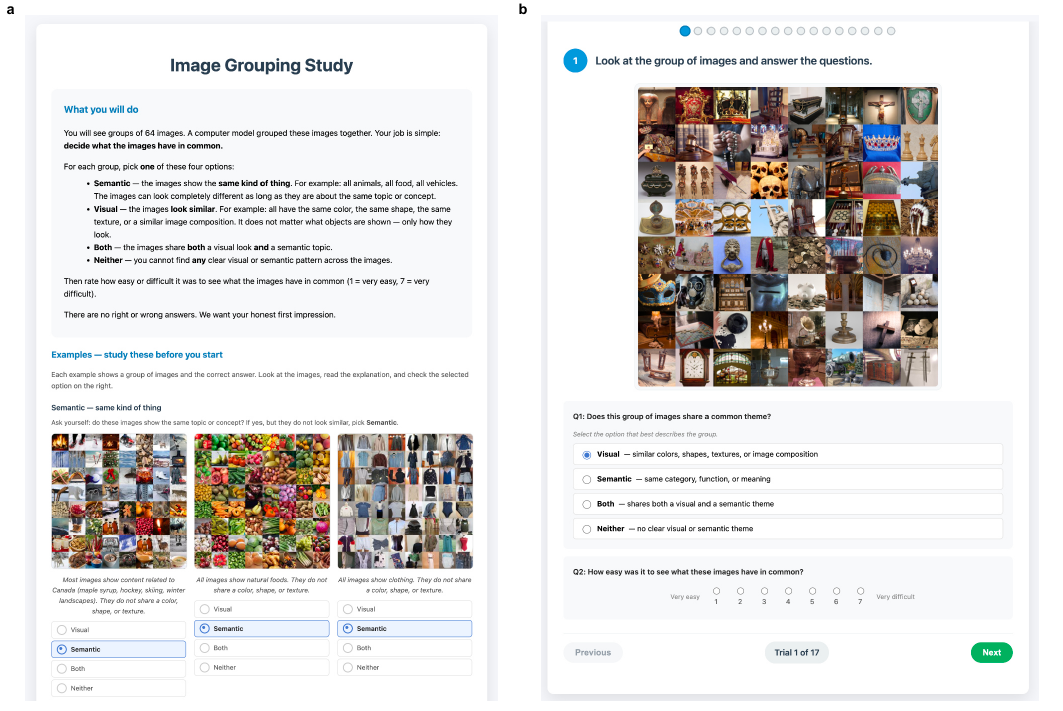}
    \caption{\textbf{Experiment instructions and task interface.} \textbf{(a)}~Instructions provided to participants before the task, including visual examples of semantic, visual, mix and neither. \textbf{(b)}~Task interface and questions asked.}
    \label{fig:connect_instructions}
\end{figure}

\paragraph{Experimental design.}
The 1{,}059 dimensions were distributed across 71 survey versions using a between-subjects design, with 15 real trials per version. Each version additionally contained one catch trial (a grid of dog images, expected response: \emph{semantic} or \emph{both}), one attention-check trial (explicit instruction: select \emph{both} and difficulty $= 4$), and one repeated real trial (the first real trial, re-shown at a random position, used to estimate within-participant consistency). Trial order within each version was randomized. Each version was assigned to 8 participants, initially yielding 8 assigned ratings per dimension before quality-control exclusions. The study was administered online via the Connect platform (see Fig.~\ref{fig:connect_instructions}). Participants were compensated at \texteuro{}0.75 per task (approximately 6 minutes per session, or \texteuro{}7.50/h).

\paragraph{Quality control.}
Participants who failed the attention-check trial (i.e.\ did not select \emph{both} with difficulty $= 4$) were excluded. Of 570 participants, 432 (76\%) passed this criterion and were retained for analysis.

\paragraph{Reliability by content type.}
Overall, raters agreed with the majority-vote label on 65\% of trials on average (per-rater mean accuracy = 0.65, median = 0.67), and majority-vote labels were stable across independent half-samples of raters, with the two halves agreeing on the majority label in 57\% of splits on average (per-dimension median split-half reliability = 0.57, mean = 0.55). Both measures varied systematically across content categories. Per-rater accuracy was highest for \emph{neither} dimensions ($\mu = 0.72$) and lowest for \emph{both} dimensions ($\mu = 0.57$), with \emph{semantic} and \emph{visual} intermediate ($\mu \approx 0.63$; Kruskal--Wallis $H = 14.6$, $p = .002$). Split-half reliability followed the same ordering: \emph{neither} dimensions yielded the most stable majority-vote labels (median = 0.77), while \emph{both} dimensions were least stable (median = 0.35), with \emph{visual} (median = 0.49) and \emph{semantic} (median = 0.43) intermediate. This convergence across two independent reliability measures confirms that the observed pattern reflects genuine differences in label-boundary ambiguity rather than artifacts of any single metric. The structure of disagreements further validates the label space. When raters disagreed on a \emph{both} dimension, they split nearly evenly between \emph{semantic} (44\%) and \emph{visual} (42\%), confirming that raters are decomposing a genuinely mixed-content dimension rather than responding randomly. When raters disagreed on a \emph{visual} dimension, they most often substituted \emph{neither} (43\%), suggesting that the visual/neither boundary is the fuzziest in the label space -- consistent with the observation that low-universality dimensions often encode subtle textural properties that can appear uninterpretable. Together, these patterns indicate that disagreements are structured and interpretable rather than random, supporting the validity of the majority-vote labels used throughout.
\FloatBarrier

\section{Universality across network depth}
\label{app:depth}

Penultimate layers often encode representations that support categorical discrimination, so the finding that shared dimensions tend to be semantic could partially reflect a layer-selection artifact rather than a property of universality itself. However, semantic content varies across dimensions within the penultimate layer, with only 19\% labeled as semantic by raters. To test whether this variation depends on the layer we chose, we repeated our pipeline at three additional depths. For each of the 162 models, we extracted representations at normalized depths of 0.1 (early), 0.5 (middle), and 0.9 (late), where depth measures the relative position among a model's processing blocks, with 0 denoting the input and 1 the penultimate layer. This normalization is a heuristic, but is necessary in light of the fact that the number of layers varies widely across models. We applied our factorization procedure and scored each dimension against the dimensions of all other models at the same depth. We then repeated the labeling experiment with the identical task, instructions, and exclusion criteria as the original study (Appendix~\ref{app:dimension_ratings}), collecting labels for 2{,}098 representative dimensions (453 early, 663 middle, 982 late; at least five raters each).

\begin{figure}[!htbp]
    \centering
    \includegraphics[width=\linewidth]{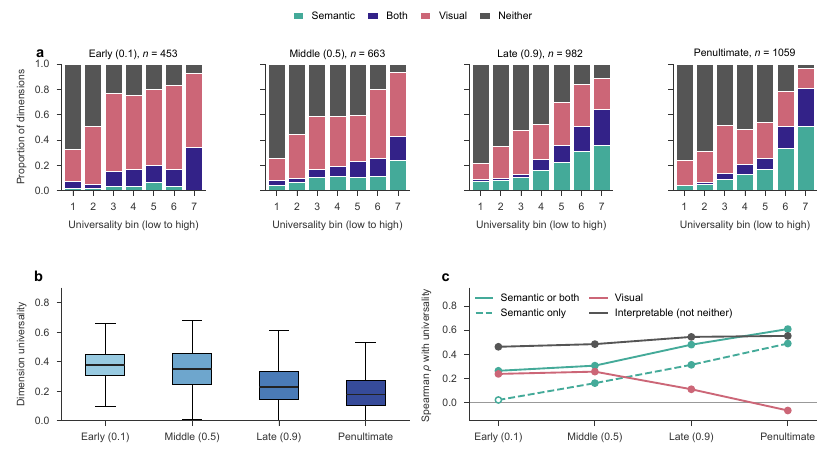}
    \caption{\textbf{Dimension content across depth.} \textbf{(a)}~Majority labels across universality bins at each depth. \textbf{(b)}~Dimension-level universality per depth, scored against the models at the same depth. \textbf{(c)}~Spearman correlation between universality and the share of raters choosing each label; open markers, $p \geq .05$.}
    \label{fig:depth_content}
\end{figure}

Absolute universality decreases with depth (model-level means: early $0.377$, middle $0.349$, late $0.243$, penultimate $0.196$; Fig.~\ref{fig:depth_content}b). However, universality is computed separately within each depth, so these values reflect agreement relative to different reference sets.

The semantic content of universal dimensions increases with depth. At the early depth, universality correlates with the share of mixed visual-semantic labels ($\rho = 0.29$), but not with the share of purely semantic labels ($\rho = 0.02$, $p = 0.62$). Raters labeled only 3\% of early dimensions as semantic and 14\% as mixed visual-semantic. To measure the tendency toward semantic content, we combined the shares of raters choosing \textit{semantic} or \textit{both} and correlated this measure with universality across rated dimensions within each depth. More universal dimensions are more semantic at every depth, and this relationship strengthens with depth (early $\rho = 0.26$, middle $0.31$, late $0.48$, penultimate $0.61$; all $p < .001$; Fig.~\ref{fig:depth_content}c).

Universal dimensions are also more interpretable than model-specific dimensions at all depths. The share of raters who judged a dimension interpretable correlates strongly with universality in every cohort (early $\rho = 0.46$, middle $0.49$, late $0.55$, penultimate $0.55$; all $p < .001$; Fig.~\ref{fig:depth_content}c). The relationship between universality and semantic content thus strengthens with depth, while the relationship with interpretability holds across the hierarchy.

\begin{figure}[!htbp]
    \centering
    \includegraphics[width=\linewidth]{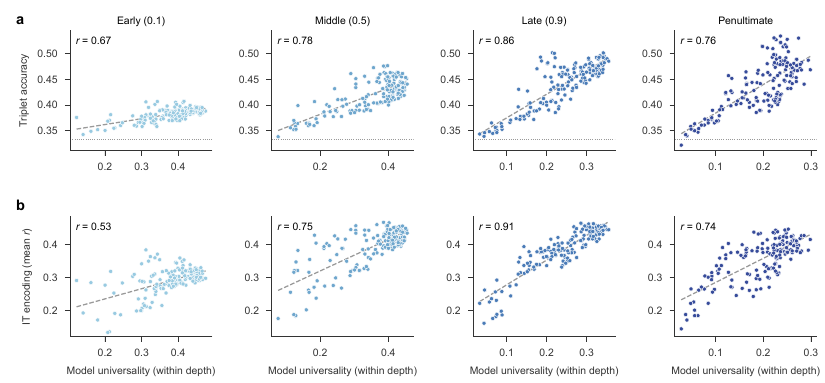}
    \caption{\textbf{Universality predicts behavior and IT activity at every depth.} Model universality vs.\ \textbf{(a)}~triplet accuracy and \textbf{(b)}~IT encoding at each depth. Dotted line: chance (1/3).}
    \label{fig:depth_prediction}
\end{figure}

Universality correlates with human odd-one-out accuracy (early $r = 0.67$, middle $0.78$, late $0.86$, penultimate $0.76$) and macaque IT encoding performance (early $r = 0.53$, middle $0.75$, late $0.91$, penultimate $0.74$; all $p < .001$; Fig.~\ref{fig:depth_prediction}). The strongest correlation with IT encoding occurs at the late depth rather than the penultimate layer. Universality is therefore informative about behavior and neural responses throughout the sampled hierarchy, rather than only close to the output.

\FloatBarrier

\section{Validating dimension labels with feature models}
\label{app:label_grounding}

To validate the human labels against measures that involve no human judgment, we predicted each dimension's loadings from two families of feature models, semantic and visual, with cross-validated ridge regression and $R^2$ as the measure of predictive accuracy. The semantic feature models are BERT embeddings of the 1{,}854 THINGS category names \citep{Devlin2019} and the eigencomponents of a WordNet Wu--Palmer similarity matrix between the category synsets \citep{Wu1994, Miller1995}. They provide a single feature vector per category and cannot distinguish the 12 exemplars of a category, so they predict category-averaged loadings. The visual feature models are 18 image statistics covering luminance, contrast, gradients, edges, color, and spatial frequency, together with the first-layer features of an untrained ResNet-50 \citep{He2016}; these differ between exemplars and predict the loadings of all 22{,}248 images directly. In both cases we cross-validate over categories, so that all exemplars of a test category are held out together and accuracy reflects generalization to unseen categories.

\begin{figure}[!htbp]
    \centering
    \includegraphics[width=\linewidth]{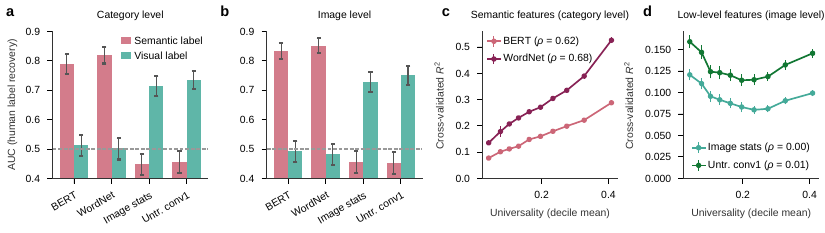}
    \caption{\textbf{Feature models validate the human labels.} \textbf{(a, b)}~AUC with which each feature model's $R^2$ identifies dimensions labeled \textit{semantic} or \textit{both} (rose) and \textit{visual} or \textit{both} (teal), at the category (a) and image level (b); dashed line, chance. \textbf{(c, d)}~$R^2$ of the semantic (c, category level) and visual (d, image level) feature models across universality deciles.}
    \label{fig:label_grounding}
\end{figure}

Dimensions that raters labeled \textit{semantic} or \textit{both} are well predicted by the semantic feature models (AUC $0.79$, 95\% CI [0.76, 0.82] for BERT; $0.82$ [0.79, 0.85] for WordNet) but not by the visual ones ($0.45$, at chance). Dimensions labeled \textit{visual} or \textit{both} show the reverse pattern, predicted by the image statistics ($0.73$ [0.69, 0.76]) and the untrained network ($0.75$ [0.72, 0.78]) while the semantic models remain at chance ($0.49$--$0.51$; Fig.~\ref{fig:label_grounding}a,b). Among dimensions labeled purely semantic or purely visual, ranking by semantic minus visual $R^2$ recovers the raters' choice between the two labels (AUC $0.90$ [0.87, 0.93]). Beyond validating the labels, the feature models recover the paper's central result without any use of the human ratings. Across all 8{,}100 dimensions, universality is predicted by a dimension's semantic $R^2$ ($\rho = 0.62$ for BERT, $0.68$ for WordNet, both $p < .001$) but not by its visual $R^2$ ($0.00$ and $0.01$; Fig.~\ref{fig:label_grounding}c,d). That universal dimensions carry categorical semantic information rather than low-level visual information can therefore be inferred from word embeddings, the WordNet taxonomy, and image statistics alone.

\FloatBarrier

\section{LLM-based dimension descriptions}
\label{app:llm_descriptions}

\begin{figure}[!htbp]
    \centering
    \includegraphics[width=\linewidth]{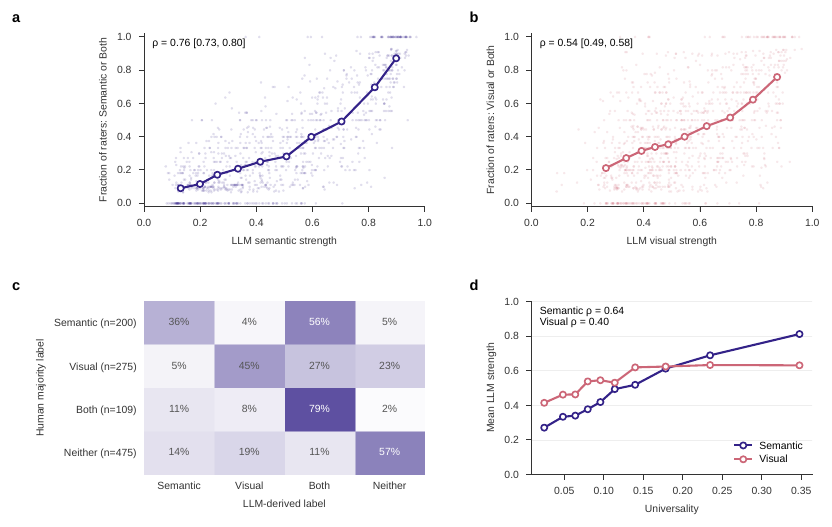}
    \caption{\textbf{LLM-based coding of dimension descriptions.} \textbf{(a)}~LLM semantic score vs.\ fraction of raters choosing \textit{semantic} or \textit{both}; \textbf{(b)}~same for visual. Circles: means of ten bins. \textbf{(c)}~LLM labels within each human label (row \%). \textbf{(d)}~Mean LLM scores across universality bins.}
    \label{fig:llm_validation}
\end{figure}

\begin{figure}[!htbp]
    \centering
    \scriptsize
    \begin{minipage}[c]{0.17\linewidth}
      \includegraphics[width=\linewidth]{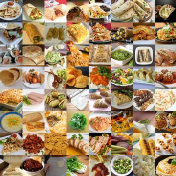}
    \end{minipage}\hfill
    \begin{minipage}[c]{0.81\linewidth}
      \textbf{Human: semantic} (8 of 9 raters)\quad universality $0.30$\quad LLM: semantic $0.89$, visual $0.49$ $\rightarrow$ \emph{semantic}\par\smallskip
      \textit{claude-sonnet-5:} A grid of close-up photographs of various prepared foods, many served on white plates or in bowls, including wraps, dumplings, salads, and other dishes shot from above or at an angle.\par
      \textit{gemini-3.6-flash:} A diverse collection of close-up food photography showcasing various cooked dishes, appetizers, snacks, and ingredients.\par
      \textit{gpt-5.6-luna:} The grid is a varied collection of close-up food photographs, with a substantial subset showing wrapped or folded foods.\par
    \end{minipage}
    \par\medskip
    \begin{minipage}[c]{0.17\linewidth}
      \includegraphics[width=\linewidth]{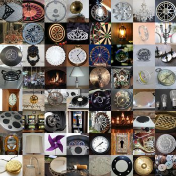}
    \end{minipage}\hfill
    \begin{minipage}[c]{0.81\linewidth}
      \textbf{Human: visual} (8 of 11 raters)\quad universality $0.15$\quad LLM: semantic $0.21$, visual $0.90$ $\rightarrow$ \emph{visual}\par\smallskip
      \textit{claude-sonnet-5:} A grid dominated by round, dial-like or wheel-like objects, many with metallic or antique finishes, photographed as isolated close-ups against plain backgrounds.\par
      \textit{gemini-3.6-flash:} A grid composed almost entirely of close-up photographs of circular or round objects, often featuring radial patterns, dials, wheels, or metallic finishes.\par
      \textit{gpt-5.6-luna:} The grid is dominated by prominently photographed, often circular objects, with frequent reflective or polished materials and relatively centered compositions.\par
    \end{minipage}
    \par\medskip
    \begin{minipage}[c]{0.17\linewidth}
      \includegraphics[width=\linewidth]{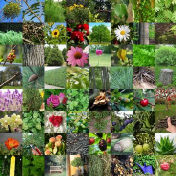}
    \end{minipage}\hfill
    \begin{minipage}[c]{0.81\linewidth}
      \textbf{Human: both} (10 of 10 raters)\quad universality $0.29$\quad LLM: semantic $0.92$, visual $0.69$ $\rightarrow$ \emph{both}\par\smallskip
      \textit{claude-sonnet-5:} A grid of close-up outdoor photos featuring plants, trees, flowers, and fruits, mostly dominated by green tones.\par
      \textit{gemini-3.6-flash:} A collection of images centered around plants, flora, foliage, flowers, tree bark, and outdoor botanical elements.\par
      \textit{gpt-5.6-luna:} The grid is dominated by close-up outdoor photographs of vegetation and individual plant features, with green foliage recurring throughout.\par
    \end{minipage}
    \par\medskip
    \begin{minipage}[c]{0.17\linewidth}
      \includegraphics[width=\linewidth]{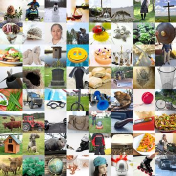}
    \end{minipage}\hfill
    \begin{minipage}[c]{0.81\linewidth}
      \textbf{Human: neither} (10 of 10 raters)\quad universality $0.01$\quad LLM: semantic $0.09$, visual $0.29$ $\rightarrow$ \emph{neither}\par\smallskip
      \textit{claude-sonnet-5:} A broad, miscellaneous collection of stock-photo style images depicting animals, food, tools, people, and objects with no single consistent theme connecting them.\par
      \textit{gemini-3.6-flash:} The image grid displays a wide variety of unrelated subjects, including animals, foods, vehicles, objects, people, and landscapes, with no single unifying theme, visual style, or subject matter across the set.\par
      \textit{gpt-5.6-luna:} The grid is visually varied, but many images emphasize a single colorful, easily recognizable subject with relatively little background clutter.\par
    \end{minipage}
    \caption{\textbf{Example LLM descriptions.} One dimension per human label: image grid, human votes, universality, averaged LLM scores and label, and each describer's summary sentence.}
    \label{fig:llm_examples}
\end{figure}

Open-ended descriptions complement the four-way labels, which target the broad typology humans apply to dimensions rather than their fine-grained content. Three multimodal LLMs (Gemini 3.6 Flash, Claude Sonnet 5, GPT-5.6 Luna) independently described each of the 1{,}059 rated dimensions from the $8 \times 8$ image grid shown to raters. Two further LLMs (Claude Sonnet 5, GPT-5.6 Terra) extracted from these descriptions continuous semantic and visual scores and specific properties including object category, function, scene context, color, shape, texture, and background. After averaging the two classifiers' scores, we thresholded each at 0.5, labeling dimensions that exceed only the semantic threshold \textit{semantic}, only the visual threshold \textit{visual}, both \textit{both}, and neither \textit{neither}. These labels agree with the human majority label for 52\% of dimensions (Fig.~\ref{fig:llm_validation}c), so description-based coding is informative but cannot simply replace direct human judgments. The continuous scores nonetheless track the raters. LLM semantic scores correlate with the proportion of raters who detected any semantic regularity ($\rho = 0.76$) and visual scores with the proportion who detected any visual regularity ($\rho = 0.54$; Fig.~\ref{fig:llm_validation}a,b), with votes of \textit{both} counting toward both measures. The descriptions also reproduce the relationship with universality, which correlates more strongly with semantic ($\rho = 0.64$) than with visual evidence ($\rho = 0.40$; Fig.~\ref{fig:llm_validation}d). Figure~\ref{fig:llm_examples} shows example descriptions.

\FloatBarrier

\section{Category consistency}
\label{app:category_consistency}

Our universality metric quantifies how consistently a dimension recurs across models, but does not reveal what a dimension represents. To characterize the content of each dimension, we exploit the categorical structure of the THINGS image set. The $N = 22{,}248$ images are organized into $C = 1{,}854$ object categories with $J = 12$ exemplars each ($N = C \times J$). Each NMF dimension $\vct{w}_{m,k} \in \mathbb{R}_{\geq 0}^{N}$ assigns a nonnegative loading to every image. If a dimension captures a semantic category, exemplars of the same category should receive similar loadings. If it instead reflects a visual property such as texture or color, exemplars within a category may receive very different loadings. We formalize this intuition using one-way ANOVA to decompose each dimension's loading variance into between- and within-category components.

For a given dimension $k$ of model $m$, let $w_{m,k,cj}$ denote the loading of the $j$-th exemplar in category $c$, let $\bar{w}_{m,k,c} = \frac{1}{J}\sum_{j=1}^{J} w_{m,k,cj}$ be the mean loading for category $c$, and let $\bar{w}_{m,k} = \frac{1}{N}\sum_{c,j} w_{m,k,cj}$ be the grand mean over all images. The total sum of squares decomposes as
\begin{equation}
  \underbrace{\sum_{c=1}^{C}\sum_{j=1}^{J}(w_{m,k,cj} - \bar{w}_{m,k})^2}_{\mathrm{SS}_{\mathrm{total}}}
  \;=\;
  \underbrace{\sum_{c=1}^{C} J\,(\bar{w}_{m,k,c} - \bar{w}_{m,k})^2}_{\mathrm{SS}_{\mathrm{between}}}
  \;+\;
  \underbrace{\sum_{c=1}^{C}\sum_{j=1}^{J}(w_{m,k,cj} - \bar{w}_{m,k,c})^2}_{\mathrm{SS}_{\mathrm{within}}}.
\end{equation}
Category consistency is the proportion of variance explained by category membership,
\begin{equation}
  \eta^2_{m,k} = \frac{\mathrm{SS}_{\mathrm{between}}}{\mathrm{SS}_{\mathrm{total}}} \;\in\; [0, 1].
\end{equation}
A dimension with $\eta^2_{m,k} \approx 1$ assigns nearly identical loadings to all exemplars of the same category and different loadings across categories. A dimension with $\eta^2_{m,k}$ near the chance level of $C / N \approx 0.08$ carries no more category information than expected from a random loading vector.

We compute $\eta^2_{m,k}$ independently for each dimension $k$ in each model $m$, yielding $r$ scores per model. To quantify reconstruction importance, we measure the drop in explained variance $\Delta R^2_{m,k}$ when dimension $k$ is removed from the low-rank reconstruction $\mat{W}_m\mat{W}_m^\top$.

\FloatBarrier
\subsection{Controlling for category consistency}
\label{app:category_control}

We tested whether category consistency explains the relationship between universality and IT or behavioral predictivity at the model level. For each model, we evaluated its mean universality, its mean category consistency ($\eta^2$), and its IT encoding performance or human odd-one-out accuracy. We then regressed category consistency out of both universality and the predictivity measure and correlated the two residuals across models.

\begin{figure}[!htbp]
    \centering
    \includegraphics[width=\linewidth]{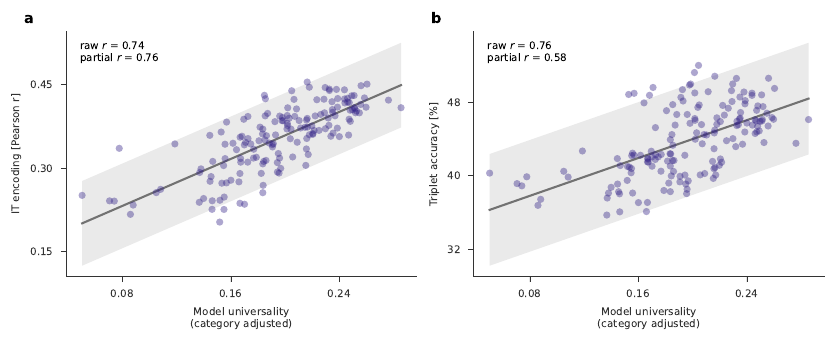}
    \caption{\textbf{Controlling for category consistency.} Model universality vs.\ \textbf{(a)}~IT encoding and \textbf{(b)}~triplet accuracy, both adjusted for each model's mean category consistency ($n = 162$). Lines: least-squares fits; bands: $\pm 1.96$ residual SD.}
    \label{fig:category_control}
\end{figure}

Controlling for category consistency leaves the relationship with macaque IT encoding essentially unchanged (raw $r = 0.743$, partial $r = 0.760$, 95\% bootstrap CI [0.697, 0.814]). The relationship with human odd-one-out accuracy weakens but stays substantial (raw $r = 0.760$, partial $r = 0.581$, 95\% bootstrap CI [0.491, 0.662]; both partial-correlation $p < .001$; Fig.~\ref{fig:category_control}). Both results hold with rank-based partial correlations (IT $\rho = 0.762$, behavior $\rho = 0.648$) and when category consistency is weighted by dimension loading importance (IT $r = 0.764$, behavior $r = 0.592$). Thus, controlling for category consistency reduces the behavioral association, whereas the IT association remains essentially unchanged.

Linearly adjusting for model-level category consistency does not remove category structure from the representations themselves. Estimating both universality and IT predictivity within each category, however, is constrained by the 12 exemplars available per category. In a preliminary test, we restricted each model's dimensions to the 12 images from each category and recomputed universality across models. The resulting scores were almost indistinguishable from chance: randomly permuting image identities yielded similar universality scores to those obtained with the actual image identities. This suggests that the number of samples is insufficient for accurately quantifying universality within categories. Repeating this analysis would benefit from a dataset with substantially more exemplars per category.

\FloatBarrier
\FloatBarrier

\section{Training data domain}
\label{app:ipcl}

Our controlled comparisons of training data are centered on ImageNet-based training, leaving open whether training-data diversity matters more broadly when datasets differ substantially from ImageNet. To address this, we computed universality for four additional models from the IPCL family \citep{Konkle2022}, whose models and training data were designed to measure domain specificity and data diversity. We used four single-domain variants trained on VGGFace2 (faces), Places-256 (scenes), OpenImages, or ImageNet-1K, holding architecture (AlexNet-GN), objective, and readout layer fixed.

\begin{figure}[!htbp]
    \centering
    \includegraphics[width=\linewidth]{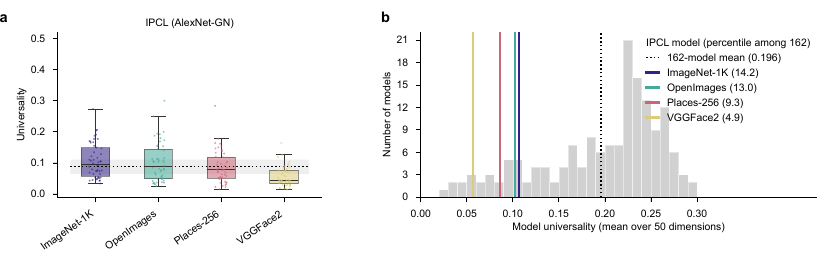}
    \caption{\textbf{Training data domain (IPCL).} \textbf{(a)}~Per-dimension universality of four AlexNet-GN IPCL models that differ only in training data. \textbf{(b)}~Model universality of the 162 reference models; colored lines mark the IPCL models (percentile in parentheses).}
    \label{fig:ipcl}
\end{figure}

The model trained on the narrowest domain has the lowest universality (VGGFace2: $U = 0.056$), and those trained on more diverse images have the highest (ImageNet-1K: $U = 0.106$; OpenImages: $U = 0.102$), with scenes in between (Places-256: $U = 0.086$; Fig.~\ref{fig:ipcl}a). The ordering is consistent with domain diversity, though the four values span a narrow range (0.050) and the two object-centric datasets are effectively tied. The pattern is therefore driven mainly by the contrast between the face-only model and the rest. All four models fall below the average universality of the full model set ($U = 0.196$), between the 4.9th and 14.2nd percentiles (Fig.~\ref{fig:ipcl}b).

Dataset size and composition covary with domain. For example, VGGFace2 is more than twice as large as ImageNet-1K but contains only faces. This comparison therefore cannot isolate a causal effect of domain diversity. The results nevertheless suggest an association between broader training domains and higher universality, consistent with findings on the model-brain alignment of these models \citep{Conwell2024}.

\FloatBarrier

\section{Neural predictivity}
\label{app:neural_predictivity}

We use multi-unit activity (MUA) recorded from inferior temporal (IT) cortex of two rhesus macaques (F and N) viewing the same 22,248 THINGS object images \citep{Papale2025}. Responses were z-scored per channel per recording day and averaged across the response time window by the original authors. Split-half reliability was pre-computed from a set of 100 test images presented with $\sim$30 repetitions each. We average reliability across splits to obtain a single reliability estimate per channel and retain only IT channels with mean reliability $> 0.3$, yielding 157 neurons for monkey F (of 320 IT channels) and 141 neurons for monkey N (of 256 IT channels).

For each of the 162 models, we fit a cross-validated ridge regression from the $r = 50$ NMF dimensions ($\mat{W}_m \in \mathbb{R}^{N \times r}$) to the response of each neuron independently. We use 5-fold cross-validation with ridge regularization selected within each training fold from $\alpha \in \{10^{-2}, \ldots, 10^{6}\}$ (20 log-spaced values). Encoding performance is the Pearson correlation between predicted and observed responses on held-out images, averaged across neurons and then across both monkeys to obtain a single encoding score per model. For the universal/specific-half comparison, we zeroed out the complementary 25 dimensions and reran the same cross-validated ridge procedure for each masked embedding.

We estimate a per-neuron noise ceiling as $\sqrt{\mathrm{reliab}_j}$, where $\mathrm{reliab}_j$ is the split-half reliability of neuron $j$ \citep{Schoppe2016}. Median noise ceilings are 0.71 (monkey F) and 0.83 (monkey N).
\FloatBarrier

\section{Behavioral predictivity}
\label{app:behavioral_predictivity}

\paragraph{Dataset.}
We use the THINGS odd-one-out triplet judgments of \citet{Hebart2020}, collected on the 1{,}854 THINGS object categories. On each trial, a participant is shown three object images and chooses the one least similar to the other two (the odd-one-out). We use the public train and validation splits, yielding $4{,}120{,}663$ train and $457{,}430$ validation triplets ($4{,}578{,}093$ total). For the zero-shot analyses reported in the main text we pool train and validation triplets into a single evaluation set, since no parameters are fit to the triplets.

\paragraph{Image selection.}
For each of the $1{,}854$ categories we use the single image on which the human behavioral ratings of \citet{Hebart2020} were collected, yielding a category-level embedding $\mat{W}_m^{\mathrm{cat}} \in \mathbb{R}^{1854 \times r}_{\geq 0}$ with $r = 50$.

\paragraph{Triplet accuracy.}
Let $\mathcal{T}$ denote the set of triplets, where each triplet is indexed by the triple of images $(i, j, k)$ for which the participant selected $k$ as the odd one out, so that $(i, j)$ is the human-chosen similar pair. For model $m$, let $s^m_{ab} = \cos(\vct{w}^m_a, \vct{w}^m_b)$ denote the cosine similarity between rows $a$ and $b$ of the category-level embedding $\mat{W}_m^{\mathrm{cat}}$. The model's predicted similar pair is
\begin{equation}
(a^{\star}, b^{\star}) \;=\; \argmax_{(a, b) \in \{(i,j),\,(i,k),\,(j,k)\}} s^m_{ab},
\end{equation}
and triplet accuracy is the fraction of triplets on which the predicted and human-chosen similar pairs agree
\begin{equation}
\mathrm{acc}(m) \;=\; \frac{1}{|\mathcal{T}|} \sum_{(i,j,k) \in \mathcal{T}} \mathbf{1}\!\left[(a^{\star}, b^{\star}) = (i, j)\right].
\end{equation}

\paragraph{Universal vs.\ specific half.}
To test whether universal dimensions drive behavioral alignment, we split each model's $r = 50$ dimensions at the median of the per-dimension universality score (Section~\ref{sec:universality}) into a universal half (top 25 dimensions) and a model-specific half (bottom 25). For each half, we construct a masked embedding by zeroing out the complementary 25 columns of $\mat{W}_m^{\mathrm{cat}}$, recompute cosine similarities on the masked embedding, and evaluate triplet accuracy exactly as above. Significance is assessed with a paired $t$-test over the 162 models on the difference $\mathrm{acc}_{\text{universal}} - \mathrm{acc}_{\text{specific}}$.
\FloatBarrier

\section{Human-aligned post-training}
\label{app:alignet}

\begin{figure}[!htbp]
    \centering
    \includegraphics[width=\linewidth]{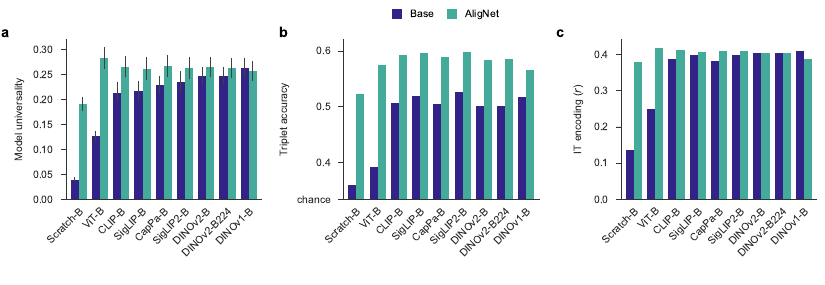}
    \caption{\textbf{Human-aligned post-training (AligNet).} \textbf{(a)}~Model universality of base and AligNet models (mean over five factorization seeds; error bars, 95\% bootstrap CIs over reference models). \textbf{(b)}~Triplet accuracy; the axis starts at chance. \textbf{(c)}~IT encoding.}
    \label{fig:alignet}
\end{figure}

Our main analyses show that universality covaries with human alignment across models, but not what needs to change in a model for its universality to increase. To test this directly, we compare nine AligNet models \citep{Muttenthaler2025}, post-trained to be more human-aligned through knowledge distillation from a surrogate teacher that was itself aligned to human similarity judgments, with their base models. Post-training increases universality in eight of nine models, from $U = 0.203$ to $0.258$ on average ($\beta = 0.055$, 95\% CI [0.011, 0.100], $p = .021$; Fig.~\ref{fig:alignet}a). Aligned models also predict triplet choices (accuracy $0.579$ vs.\ $0.480$) and IT activity (mean $r = 0.404$ vs.\ $0.352$) better than their base models (Fig.~\ref{fig:alignet}b,c). Universality can thus be increased by aligning a model's representations with human behavior, and the aligned models are at the same time better predictors of behavior and IT, supporting our main analyses.

\FloatBarrier

\section{Compute resources}
\label{app:compute}

All experiments were run on an internal high-performance computing cluster. Feature extraction from 162 models on THINGS and ObjectNet images, at three additional depths, and for the untrained, IPCL, and AligNet models required approximately 20 GPU-hours on NVIDIA A100 40GB GPUs. All factorizations required approximately 88,000 CPU-hours on CPU compute nodes with 512 GB RAM. Beyond the reported experiments, preliminary and exploratory runs required approximately an additional 5,000 CPU-hours.
\FloatBarrier

\clearpage
\section*{NeurIPS Paper Checklist}

\begin{enumerate}

\item {\bf Claims}
    \item[] Question: Do the main claims made in the abstract and introduction accurately reflect the paper's contributions and scope?
    \item[] Answer: \answerYes{} 
    \item[] Justification: The abstract and introduction state four contributions: vision models share universal dimensions, universal dimensions reflect semantic properties, no single inductive bias explains universality, and universality predicts primate neural and human behavioral alignment. Each is directly supported by analysis results. 
    \item[] Guidelines:
    \begin{itemize}
        \item The answer \answerNA{} means that the abstract and introduction do not include the claims made in the paper.
        \item The abstract and/or introduction should clearly state the claims made, including the contributions made in the paper and important assumptions and limitations. A \answerNo{} or \answerNA{} answer to this question will not be perceived well by the reviewers. 
        \item The claims made should match theoretical and experimental results, and reflect how much the results can be expected to generalize to other settings. 
        \item It is fine to include aspirational goals as motivation as long as it is clear that these goals are not attained by the paper. 
    \end{itemize}

\item {\bf Limitations}
    \item[] Question: Does the paper discuss the limitations of the work performed by the authors?
    \item[] Answer: \answerYes{} 
    \item[] Justification: Section~\ref{sec:limitations} discusses limitations of the model set, the additive non-negative decomposition, and the choice of penultimate-layer representations.
    \item[] Guidelines:
    \begin{itemize}
        \item The answer \answerNA{} means that the paper has no limitation while the answer \answerNo{} means that the paper has limitations, but those are not discussed in the paper. 
        \item The authors are encouraged to create a separate ``Limitations'' section in their paper.
        \item The paper should point out any strong assumptions and how robust the results are to violations of these assumptions (e.g., independence assumptions, noiseless settings, model well-specification, asymptotic approximations only holding locally). The authors should reflect on how these assumptions might be violated in practice and what the implications would be.
        \item The authors should reflect on the scope of the claims made, e.g., if the approach was only tested on a few datasets or with a few runs. In general, empirical results often depend on implicit assumptions, which should be articulated.
        \item The authors should reflect on the factors that influence the performance of the approach. For example, a facial recognition algorithm may perform poorly when image resolution is low or images are taken in low lighting. Or a speech-to-text system might not be used reliably to provide closed captions for online lectures because it fails to handle technical jargon.
        \item The authors should discuss the computational efficiency of the proposed algorithms and how they scale with dataset size.
        \item If applicable, the authors should discuss possible limitations of their approach to address problems of privacy and fairness.
        \item While the authors might fear that complete honesty about limitations might be used by reviewers as grounds for rejection, a worse outcome might be that reviewers discover limitations that aren't acknowledged in the paper. The authors should use their best judgment and recognize that individual actions in favor of transparency play an important role in developing norms that preserve the integrity of the community. Reviewers will be specifically instructed to not penalize honesty concerning limitations.
    \end{itemize}

\item {\bf Theory assumptions and proofs}
    \item[] Question: For each theoretical result, does the paper provide the full set of assumptions and a complete (and correct) proof?
    \item[] Answer: \answerNA{} 
    \item[] Justification: The paper does not include theoretical results.
    \item[] Guidelines:
    \begin{itemize}
        \item The answer \answerNA{} means that the paper does not include theoretical results. 
        \item All the theorems, formulas, and proofs in the paper should be numbered and cross-referenced.
        \item All assumptions should be clearly stated or referenced in the statement of any theorems.
        \item The proofs can either appear in the main paper or the supplemental material, but if they appear in the supplemental material, the authors are encouraged to provide a short proof sketch to provide intuition. 
        \item Inversely, any informal proof provided in the core of the paper should be complemented by formal proofs provided in appendix or supplemental material.
        \item Theorems and Lemmas that the proof relies upon should be properly referenced. 
    \end{itemize}

    \item {\bf Experimental result reproducibility}
    \item[] Question: Does the paper fully disclose all the information needed to reproduce the main experimental results of the paper to the extent that it affects the main claims and/or conclusions of the paper (regardless of whether the code and data are provided or not)?
    \item[] Answer: \answerYes{} 
    \item[] Justification: The Methods section and Appendices~\ref{app:models}--\ref{app:neural_predictivity} fully specify the dataset, model set, factorization protocol, universality metric, and ridge regression details.
    \item[] Guidelines:
    \begin{itemize}
        \item The answer \answerNA{} means that the paper does not include experiments.
        \item If the paper includes experiments, a \answerNo{} answer to this question will not be perceived well by the reviewers: Making the paper reproducible is important, regardless of whether the code and data are provided or not.
        \item If the contribution is a dataset and\slash or model, the authors should describe the steps taken to make their results reproducible or verifiable. 
        \item Depending on the contribution, reproducibility can be accomplished in various ways. For example, if the contribution is a novel architecture, describing the architecture fully might suffice, or if the contribution is a specific model and empirical evaluation, it may be necessary to either make it possible for others to replicate the model with the same dataset, or provide access to the model. In general. releasing code and data is often one good way to accomplish this, but reproducibility can also be provided via detailed instructions for how to replicate the results, access to a hosted model (e.g., in the case of a large language model), releasing of a model checkpoint, or other means that are appropriate to the research performed.
        \item While NeurIPS does not require releasing code, the conference does require all submissions to provide some reasonable avenue for reproducibility, which may depend on the nature of the contribution. For example
        \begin{enumerate}
            \item If the contribution is primarily a new algorithm, the paper should make it clear how to reproduce that algorithm.
            \item If the contribution is primarily a new model architecture, the paper should describe the architecture clearly and fully.
            \item If the contribution is a new model (e.g., a large language model), then there should either be a way to access this model for reproducing the results or a way to reproduce the model (e.g., with an open-source dataset or instructions for how to construct the dataset).
            \item We recognize that reproducibility may be tricky in some cases, in which case authors are welcome to describe the particular way they provide for reproducibility. In the case of closed-source models, it may be that access to the model is limited in some way (e.g., to registered users), but it should be possible for other researchers to have some path to reproducing or verifying the results.
        \end{enumerate}
    \end{itemize}

\item {\bf Open access to data and code}
    \item[] Question: Does the paper provide open access to the data and code, with sufficient instructions to faithfully reproduce the main experimental results, as described in supplemental material?
    \item[] Answer: \answerYes{} 
    \item[] Justification: Code and installation instructions are available at \url{https://github.com/florianmahner/universal-representations}. The selected rank-50 embeddings for all 162 models on THINGS and ObjectNet, with image-order metadata and bandwidth and seed selections, are available at \url{https://doi.org/10.5281/zenodo.23979573}. The Methods section and appendices describe the analysis procedures.
    \item[] Guidelines:
    \begin{itemize}
        \item The answer \answerNA{} means that paper does not include experiments requiring code.
        \item Please see the NeurIPS code and data submission guidelines (\url{https://neurips.cc/public/guides/CodeSubmissionPolicy}) for more details.
        \item While we encourage the release of code and data, we understand that this might not be possible, so \answerNo{} is an acceptable answer. Papers cannot be rejected simply for not including code, unless this is central to the contribution (e.g., for a new open-source benchmark).
        \item The instructions should contain the exact command and environment needed to run to reproduce the results. See the NeurIPS code and data submission guidelines (\url{https://neurips.cc/public/guides/CodeSubmissionPolicy}) for more details.
        \item The authors should provide instructions on data access and preparation, including how to access the raw data, preprocessed data, intermediate data, and generated data, etc.
        \item The authors should provide scripts to reproduce all experimental results for the new proposed method and baselines. If only a subset of experiments are reproducible, they should state which ones are omitted from the script and why.
        \item At submission time, to preserve anonymity, the authors should release anonymized versions (if applicable).
        \item Providing as much information as possible in supplemental material (appended to the paper) is recommended, but including URLs to data and code is permitted.
    \end{itemize}

\item {\bf Experimental setting/details}
    \item[] Question: Does the paper specify all the training and test details (e.g., data splits, hyperparameters, how they were chosen, type of optimizer) necessary to understand the results?
    \item[] Answer: \answerYes{} 
    \item[] Justification: The Methods section specifies dataset and model choices; Appendices~\ref{app:rbf}, \ref{app:optimization}, \ref{app:metric_details}, and \ref{app:neural_predictivity} give the RBF bandwidth grid, NMF optimization protocol, universality metric calibration, and ridge regression details.
    \item[] Guidelines:
    \begin{itemize}
        \item The answer \answerNA{} means that the paper does not include experiments.
        \item The experimental setting should be presented in the core of the paper to a level of detail that is necessary to appreciate the results and make sense of them.
        \item The full details can be provided either with the code, in appendix, or as supplemental material.
    \end{itemize}

\item {\bf Experiment statistical significance}
    \item[] Question: Does the paper report error bars suitably and correctly defined or other appropriate information about the statistical significance of the experiments?
    \item[] Answer: \answerYes{} 
    \item[] Justification: Controlled comparisons (Section~\ref{sec:what_drives_universality}) use Welch's $t$-test or one-way ANOVA with Bonferroni correction; bootstrap tests use $10{,}000$ resamples. The universality null uses $B=1{,}000$ row-permutations (Appendix~\ref{app:metric_details}), and error bands show 95\% bootstrap CIs (Figs.~\ref{fig:stability},~\ref{fig:sigma_selection}).
    \item[] Guidelines:
    \begin{itemize}
        \item The answer \answerNA{} means that the paper does not include experiments.
        \item The authors should answer \answerYes{} if the results are accompanied by error bars, confidence intervals, or statistical significance tests, at least for the experiments that support the main claims of the paper.
        \item The factors of variability that the error bars are capturing should be clearly stated (for example, train/test split, initialization, random drawing of some parameter, or overall run with given experimental conditions).
        \item The method for calculating the error bars should be explained (closed form formula, call to a library function, bootstrap, etc.)
        \item The assumptions made should be given (e.g., Normally distributed errors).
        \item It should be clear whether the error bar is the standard deviation or the standard error of the mean.
        \item It is OK to report 1-sigma error bars, but one should state it. The authors should preferably report a 2-sigma error bar than state that they have a 96\% CI, if the hypothesis of Normality of errors is not verified.
        \item For asymmetric distributions, the authors should be careful not to show in tables or figures symmetric error bars that would yield results that are out of range (e.g., negative error rates).
        \item If error bars are reported in tables or plots, the authors should explain in the text how they were calculated and reference the corresponding figures or tables in the text.
    \end{itemize}

\item {\bf Experiments compute resources}
    \item[] Question: For each experiment, does the paper provide sufficient information on the computer resources (type of compute workers, memory, time of execution) needed to reproduce the experiments?
    \item[] Answer: \answerYes{} 
    \item[] Justification: Compute resources are specified in Appendix ~\ref{app:compute}.
    \item[] Guidelines:
    \begin{itemize}
        \item The answer \answerNA{} means that the paper does not include experiments.
        \item The paper should indicate the type of compute workers CPU or GPU, internal cluster, or cloud provider, including relevant memory and storage.
        \item The paper should provide the amount of compute required for each of the individual experimental runs as well as estimate the total compute. 
        \item The paper should disclose whether the full research project required more compute than the experiments reported in the paper (e.g., preliminary or failed experiments that didn't make it into the paper). 
    \end{itemize}
    
\item {\bf Code of ethics}
    \item[] Question: Does the research conducted in the paper conform, in every respect, with the NeurIPS Code of Ethics \url{https://neurips.cc/public/EthicsGuidelines}?
    \item[] Answer: \answerYes{} 
    \item[] Justification: The research conforms with the NeurIPS Code of Ethics. No new pretrained models or scraped datasets are released. The crowdsourced human dimension-rating study was conducted via the CloudResearch Connect platform, with participants compensated at \texteuro{}0.75 per approximately 6-minute session (\texteuro{}7.50/h, approximately \$8/h), above both the platform's minimum pay policy of \$7.50/h and the U.S. federal minimum wage of \$7.25/h.
    \item[] Guidelines:
    \begin{itemize}
        \item The answer \answerNA{} means that the authors have not reviewed the NeurIPS Code of Ethics.
        \item If the authors answer \answerNo, they should explain the special circumstances that require a deviation from the Code of Ethics.
        \item The authors should make sure to preserve anonymity (e.g., if there is a special consideration due to laws or regulations in their jurisdiction).
    \end{itemize}

\item {\bf Broader impacts}
    \item[] Question: Does the paper discuss both potential positive societal impacts and negative societal impacts of the work performed?
    \item[] Answer: \answerNA{} 
    \item[] Justification: This paper contains foundational research about the shared representational structure across existing public vision models. There is no direct path to harmful applications.
    \item[] Guidelines:
    \begin{itemize}
        \item The answer \answerNA{} means that there is no societal impact of the work performed.
        \item If the authors answer \answerNA{} or \answerNo, they should explain why their work has no societal impact or why the paper does not address societal impact.
        \item Examples of negative societal impacts include potential malicious or unintended uses (e.g., disinformation, generating fake profiles, surveillance), fairness considerations (e.g., deployment of technologies that could make decisions that unfairly impact specific groups), privacy considerations, and security considerations.
        \item The conference expects that many papers will be foundational research and not tied to particular applications, let alone deployments. However, if there is a direct path to any negative applications, the authors should point it out. For example, it is legitimate to point out that an improvement in the quality of generative models could be used to generate Deepfakes for disinformation. On the other hand, it is not needed to point out that a generic algorithm for optimizing neural networks could enable people to train models that generate Deepfakes faster.
        \item The authors should consider possible harms that could arise when the technology is being used as intended and functioning correctly, harms that could arise when the technology is being used as intended but gives incorrect results, and harms following from (intentional or unintentional) misuse of the technology.
        \item If there are negative societal impacts, the authors could also discuss possible mitigation strategies (e.g., gated release of models, providing defenses in addition to attacks, mechanisms for monitoring misuse, mechanisms to monitor how a system learns from feedback over time, improving the efficiency and accessibility of ML).
    \end{itemize}
    
\item {\bf Safeguards}
    \item[] Question: Does the paper describe safeguards that have been put in place for responsible release of data or models that have a high risk for misuse (e.g., pre-trained language models, image generators, or scraped datasets)?
    \item[] Answer: \answerNA{} 
    \item[] Justification: The paper does not release pretrained models, generative models, or scraped datasets. All vision models and image stimuli analyzed in the paper are from existing, publicly released sources.
    \item[] Guidelines:
    \begin{itemize}
        \item The answer \answerNA{} means that the paper poses no such risks.
        \item Released models that have a high risk for misuse or dual-use should be released with necessary safeguards to allow for controlled use of the model, for example by requiring that users adhere to usage guidelines or restrictions to access the model or implementing safety filters. 
        \item Datasets that have been scraped from the Internet could pose safety risks. The authors should describe how they avoided releasing unsafe images.
        \item We recognize that providing effective safeguards is challenging, and many papers do not require this, but we encourage authors to take this into account and make a best faith effort.
    \end{itemize}

\item {\bf Licenses for existing assets}
    \item[] Question: Are the creators or original owners of assets (e.g., code, data, models), used in the paper, properly credited and are the license and terms of use explicitly mentioned and properly respected?
    \item[] Answer: \answerYes{} 
    \item[] Justification: All datasets and models used in this study are from existing, publicly released sources and are properly cited in the paper. THINGS images \citep{Hebart2019} are distributed under CC~BY~4.0. ObjectNet \citep{Barbu2019} is distributed under its research license and is used here only for feature extraction from pretrained models, not for redistribution. Pretrained model weights come from the model set of \citet{Conwell2024} (aggregated from timm, torchvision, VISSL, and Taskonomy; license terms of each source are respected), OpenCLIP \citep{Cherti2023} (MIT License), and the SLIP family \citep{Mu2022}. Human behavioral triplet data \citep{Hebart2020} and macaque IT recordings \citep{Papale2025} are used under the terms specified by the original publications.
    \item[] Guidelines:
    \begin{itemize}
        \item The answer \answerNA{} means that the paper does not use existing assets.
        \item The authors should cite the original paper that produced the code package or dataset.
        \item The authors should state which version of the asset is used and, if possible, include a URL.
        \item The name of the license (e.g., CC-BY 4.0) should be included for each asset.
        \item For scraped data from a particular source (e.g., website), the copyright and terms of service of that source should be provided.
        \item If assets are released, the license, copyright information, and terms of use in the package should be provided. For popular datasets, \url{paperswithcode.com/datasets} has curated licenses for some datasets. Their licensing guide can help determine the license of a dataset.
        \item For existing datasets that are re-packaged, both the original license and the license of the derived asset (if it has changed) should be provided.
        \item If this information is not available online, the authors are encouraged to reach out to the asset's creators.
    \end{itemize}

\item {\bf New assets}
    \item[] Question: Are new assets introduced in the paper well documented and is the documentation provided alongside the assets?
    \item[] Answer: \answerYes{} 
    \item[] Justification: We release the selected rank-50 non-negative embeddings for all 162 vision models on THINGS and ObjectNet under CC BY 4.0 at \url{https://doi.org/10.5281/zenodo.23979573}, together with image-order metadata and per-model RBF bandwidth and seed selections. The release includes a README with loading instructions and file conventions. Code and model metadata are available under BSD-3-Clause, with separate notices for third-party code at \url{https://github.com/florianmahner/universal-representations}.
    \item[] Guidelines:
    \begin{itemize}
        \item The answer \answerNA{} means that the paper does not release new assets.
        \item Researchers should communicate the details of the dataset\slash code\slash model as part of their submissions via structured templates. This includes details about training, license, limitations, etc. 
        \item The paper should discuss whether and how consent was obtained from people whose asset is used.
        \item At submission time, remember to anonymize your assets (if applicable). You can either create an anonymized URL or include an anonymized zip file.
    \end{itemize}

\item {\bf Crowdsourcing and research with human subjects}
    \item[] Question: For crowdsourcing experiments and research with human subjects, does the paper include the full text of instructions given to participants and screenshots, if applicable, as well as details about compensation (if any)? 
    \item[] Answer: \answerYes{} 
    \item[] Justification: Appendix~\ref{app:dimension_ratings} provides the full task description, screenshots of the task interface (Fig.~\ref{fig:connect_instructions}), and compensation details.
    \item[] Guidelines:
    \begin{itemize}
        \item The answer \answerNA{} means that the paper does not involve crowdsourcing nor research with human subjects.
        \item Including this information in the supplemental material is fine, but if the main contribution of the paper involves human subjects, then as much detail as possible should be included in the main paper. 
        \item According to the NeurIPS Code of Ethics, workers involved in data collection, curation, or other labor should be paid at least the minimum wage in the country of the data collector. 
    \end{itemize}

\item {\bf Institutional review board (IRB) approvals or equivalent for research with human subjects}
    \item[] Question: Does the paper describe potential risks incurred by study participants, whether such risks were disclosed to the subjects, and whether Institutional Review Board (IRB) approvals (or an equivalent approval/review based on the requirements of your country or institution) were obtained?
    \item[] Answer: \answerYes{} 
    \item[] Justification: The crowdsourced human dimension-rating study was approved by the local ethics committee (reference 27/22-ek). The macaque IT recordings and human behavioral triplet data were collected under the approvals reported in their original publications \citep{Papale2025, Hebart2020}.
    \item[] Guidelines:
    \begin{itemize}
        \item The answer \answerNA{} means that the paper does not involve crowdsourcing nor research with human subjects.
        \item Depending on the country in which research is conducted, IRB approval (or equivalent) may be required for any human subjects research. If you obtained IRB approval, you should clearly state this in the paper. 
        \item We recognize that the procedures for this may vary significantly between institutions and locations, and we expect authors to adhere to the NeurIPS Code of Ethics and the guidelines for their institution. 
        \item For initial submissions, do not include any information that would break anonymity (if applicable), such as the institution conducting the review.
    \end{itemize}

\item {\bf Declaration of LLM usage}
    \item[] Question: Does the paper describe the usage of LLMs if it is an important, original, or non-standard component of the core methods in this research? Note that if the LLM is used only for writing, editing, or formatting purposes and does \emph{not} impact the core methodology, scientific rigor, or originality of the research, declaration is not required.
    \item[] Answer: \answerNA{} 
    \item[] Justification: LLMs were not used as a component of the core methods.
    \item[] Guidelines:
    \begin{itemize}
        \item The answer \answerNA{} means that the core method development in this research does not involve LLMs as any important, original, or non-standard components.
        \item Please refer to our LLM policy in the NeurIPS handbook for what should or should not be described.
    \end{itemize}

\end{enumerate}

\end{document}